\documentclass{article}

\usepackage{array}
\newcolumntype{C}[1]{>{\centering\arraybackslash}p{#1}}

 \usepackage[main, final]{neurips_2026}

\usepackage[utf8]{inputenc} 
\usepackage[T1]{fontenc}    
\usepackage{hyperref}       
\hypersetup{hidelinks}
\usepackage{url}            
\usepackage{booktabs}       
\usepackage{amsfonts}       
\usepackage{nicefrac}       
\usepackage{microtype}      
\usepackage{xcolor}         
\usepackage{graphicx}
\usepackage{amsmath}
\usepackage{fontawesome5} 

\title{Tell Me What To Learn: Generalizing Neural Memory to be Controllable in Natural Language}

\newcommand{\equalcontrib}{\textsuperscript{*}}

\author{%
  Max S. Bennett\equalcontrib \\ 
  Columbia University\\
  \texttt{max.bennett@columbia.edu} \\
  \And
  Thomas P. Zollo\equalcontrib \\
  Columbia University \\
  \texttt{tpz2105@columbia.edu} \\
  \AND
  Richard Zemel \\
  Columbia University \\
  \texttt{zemel@cs.columbia.edu} \\
}

\begin{document}

\maketitle

\begin{abstract}

Modern machine learning models are deployed in diverse, non-stationary environments where they must continually adapt to new tasks and evolving knowledge. 
Continual fine-tuning and in-context learning are costly and brittle, whereas neural memory methods promise lightweight updates with minimal forgetting. 
However, existing neural memory models assume a single fixed objective and homogeneous information streams, leaving users with no control over what the model remembers or ignores over time. 
To address this challenge, we propose a generalized neural memory system that performs flexible updates based on learning instructions specified in natural language. 
Our approach enables adaptive agents to learn selectively from heterogeneous information sources, supporting settings such as healthcare and customer service where fixed-objective memory updates are insufficient. The code and models for this paper are publicly available at:  \href{https://github.com/maxbennett/generalized-neural-memory}{\faGithub~https://github.com/maxbennett/generalized-neural-memory} 

\end{abstract}

\section{Introduction}\label{sec:introduction}

Modern foundation models, including large language models (LLMs), acquire broad skills and world knowledge during large-scale pretraining \citep{brown2020languagemodelsfewshotlearners}. 
However, real-world deployment exposes them to non-stationary environments that demand continual adaptation to new tasks and evolving knowledge \citep{lazaridou2021mindgapassessingtemporal}. 
While fine-tuning and other gradient-based techniques are effective for adaptation, they are costly, require data or environments to be available before deployment, and often fail when applied repeatedly due to catastrophic forgetting \citep{dautume2019episodicmemorylifelonglanguage}.
Retrieval-augmented generation (RAG) \citep{lewis2021retrievalaugmentedgenerationknowledgeintensivenlp} and in-context learning (ICL) \citep{brown2020languagemodelsfewshotlearners} enable more seamless adaptation.
Still, per-instance retrieval can be cumbersome and imprecise, and ICL suffers from both the quadratic cost of Transformer attention \citep{vaswani2017attention} and significant performance degradation as context length grows \citep{shi2023largelanguagemodelseasily, liu2023lostmiddlelanguagemodels}. 

\begin{figure*}[!t]
\vspace{-15pt}
	\centering
	\includegraphics[width=\textwidth]{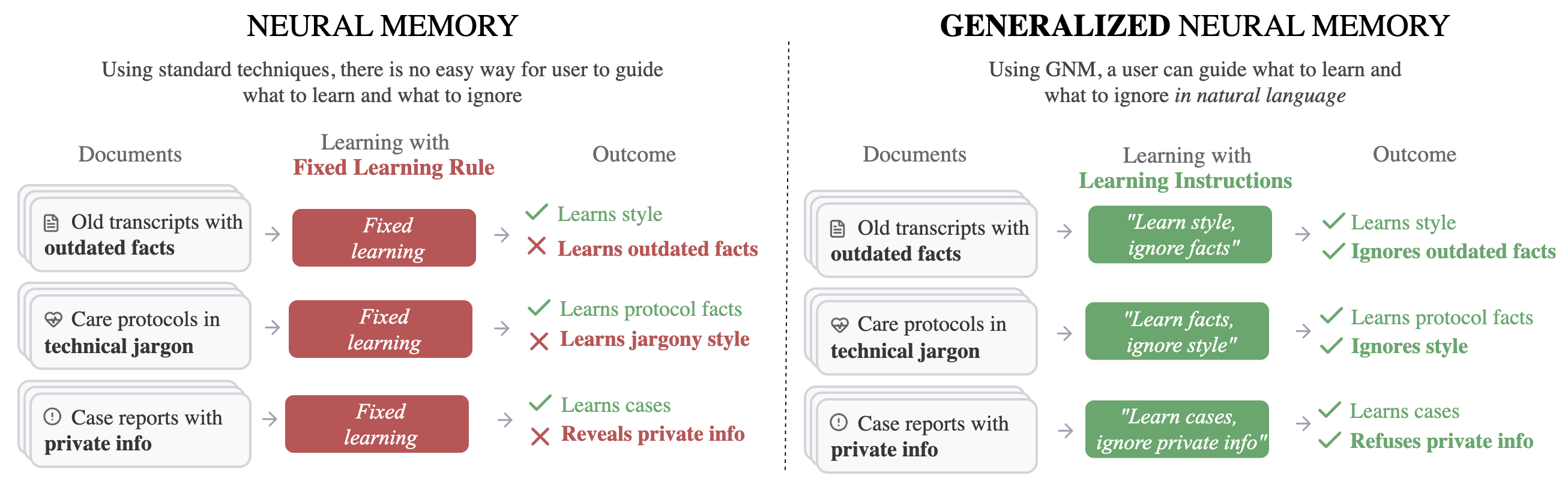}
	\caption{
    An AI system with memory can adapt to its environment continuously by integrating diverse information sources. We propose a generalized neural memory system that performs flexible long-term updates based on learning instructions specified in natural language.  Our approach enables important use cases in critical domains such as healthcare, where an adaptive agent must learn from heterogeneous documents that preclude using a neural memory system with a fixed objective.}
	\label{fig:fig1}
\vspace{-12pt}
\end{figure*}

Neural memory has emerged as a possible solution to these challenges and presents a promising middle ground \citep{sukhbaatar2015endtoend, bulatov2022rmt, behrouz_titans_2024}. 
Broadly, neural memory approaches equip a model with an external, often differentiable memory that can be updated over time and accessed at inference, allowing the model to accumulate information across interactions without retraining. 
New information is encoded into, e.g., compact vector slots or dedicated memory tokens, and the system learns how to write to and read from this memory to modulate the model’s predictions.
Neural memory mitigates the catastrophic forgetting associated with fine-tuning, because at test time the model parameters are not modified and responses change only via the memory interface.
Further, by compressing ``what to remember'' into a fixed-size neural memory, it avoids the token-consumption challenges of ICL as well as the system overhead associated with retrieval.
This approach has recently shown state-of-the-art performance on various long-context benchmarks, including language modeling, needle-in-the-haystack retrieval, time series forecasting, and DNA modeling \citep{behrouz_titans_2024, wang2024memoryllmselfupdatablelargelanguage}.

Despite this progress, neural memory systems are typically designed under a single notion of ``what to learn,'' implicitly assuming that all future experience will come from homogeneous data sources and should be learned according to a fixed objective.
In many real-world settings, however, different documents expose different aspects the system should internalize: facts, style, tone, escalation heuristics, safety constraints, or other information.
These can conflict across sources, and the same source might call for different learning goals depending on the downstream application. 
Though downstream users often know precisely which parts of which inputs should be remembered or ignored, current neural memory systems provide no direct mechanism for expressing that intent. 
For example, in a medical practice deploying an AI agent for post-operative support,
doctors may want the agent to learn from years of nurse–patient transcripts when to escalate to a human versus answer autonomously, while explicitly avoiding the outdated factual protocols mentioned in those calls. 
At the same time, they want the AI agent to absorb accurate, up-to-date procedural and billing facts from regularly updated internal protocol documents, without adopting their overly technical tone.

To address this challenge, we introduce a \textbf{\emph{generalized neural memory} (GNM)} approach that enables memory updates to be guided via natural-language instructions, effectively allowing the system to be 
told what to learn from new data.
Our instruction-conditioned memory turns a document/instruction pair into a targeted, long-term update, yielding a controllable, efficient alternative to ICL and RAG while avoiding the destructive interference and cost of 
repeated
training.
Instructions may come from end users, system designers, persistent policies, or automated components that infer the appropriate learning objective from context, enabling lightweight, selective learning of information over time.
In experiments, our approach demonstrates generalization to unseen instructions across document and preference learning tasks.
Further, our results highlight the value of the information bottleneck introduced by neural memory in this setting, as ICL and RAG approaches struggle (even when fine-tuned) because of their inability to ignore non-relevant memories.

To summarize our contributions:
\begin{itemize}
\item  We introduce language-controlled neural memory, enabling users to specify what to memorize or ignore via natural-language learning instructions.
\item We empirically demonstrate generalization to unseen instructions, outperforming in-context learning and RAG approaches on selectivity and efficiency.
\item We show through analysis and ablation interventions that our language-controlled neural memory learns to selectively modify what is memorized based on language prompts, and is not merely changing what to attend to in memory.
\end{itemize}
Our goal in this work is to move the field toward AI agents that continuously adapt as collaborative, lifelong learning partners, suitable for deployment in safety-critical, evolving domains.

\paragraph{Related Work.}
The field of continual learning studies how models can incrementally acquire new knowledge (plasticity) while retaining existing capabilities (stability) \citep{mcclelland1995there, MCCLOSKEY1989109, HADSELL20201028}.
A range of strategies have been proposed to accomplish this, including
regularization-based methods \citep{kirkpatrick2017ewc, zenke2017synaptic, ritter2018online},
replay-based approaches \citep{lopezpaz2017gem, shin2017dgr, buzzega2020dark},
architectural modifications \citep{rusu2016progressive, fernando2017pathnet, yoon2018lifelong},
and optimization-based methods \citep{farajtabar2020orthogonal, javed2019meta}. 
Despite their success in controlled benchmarks, many of these methods rely on explicit task boundaries, replay buffers, or repeated retraining, assumptions that break down in realistic settings with weak supervision, blurred task boundaries, and limited compute or storage budgets \citep{wang2024comprehensive, bang2022onlinecontinuallearningcontaminated, ren2021wanderingworldonlinecontextualized, dautume2019episodicmemorylifelonglanguage}.

Neural memory offers an approach to continual adaptation wherein compressed information can be written, stored, and retrieved without overwriting core model weights \citep{graves2014ntm, weston2014memory, miller2016kv}. 
Recent work integrates neural memory into transformers and large language models to extend effective context and support long-horizon behavior \citep{bulatov2022rmt, behrouz_titans_2024, wang2024memoryllmselfupdatablelargelanguage}. 
For example, MemoryLLM \citep{wang2024memoryllmselfupdatablelargelanguage} compresses documents into latent memory tokens used as layer-wise prefixes, while Titans \citep{behrouz_titans_2024} 
updates neural memory module weights at test time based on surprising new information.
We extend this work by generalizing neural memory beyond fixed learning objectives, enabling memory updates to be controlled by natural-language instructions.

Beyond latent neural memory, recent frameworks use LLMs to manage explicit external memory (e.g., text storage and retrieval) in long-context tasks, with some learning memory-management policies through reinforcement learning \citep{chhikara2025mem0,wang2025mem,yan2026memory}.
Related retrieval-based approaches enable non-parametric adaptation without additional training through kNN-style language models and associative key--value memory \citep{khandelwal2020knnlm,he2024camelot}, while query-aware compression \citep{jiang2024longllmlingua} selects information for a particular question.
Finally, knowledge editing techniques may be used to modify facts in LLM weights \citep{meng_locating_2023}.

\section{Generalizing Neural Memory to be Controllable with Language}\label{sec:methods}

Our goal is to enable controllable neural memory updates via natural-language learning instructions.
We consider a setting in which an LLM equipped with neural memory interacts with a stream of documents (each associated with a learning instruction) and user queries.
Memory is updated only when a new document arrives; user queries are answered using the current memory state.

Formally, we assume a language model $f_\theta$ with parameters $\theta$, augmented with a neural memory state $M \in \mathcal{M}$, where $M_t$ denotes the memory state after $t$ updates.
We write $p_\theta(y\mid \cdot,M_t)$ for the conditional distribution over responses induced by $f_\theta$ when conditioned on memory state $M_t$.
The memory $M_t$ can be implemented in numerous ways, including with continuous prefix embeddings, a bank of memory vectors, or layer-wise memory tokens maintained by a long-term memory module; our formulation is agnostic to this choice as long as $M_t$ can be updated and used to condition $f_\theta$.

Documents arrive in a sequence 
$S = ((I_t, D_t))_{t=1}^T \sim \mathcal{S}$,
where $D_t$ is a document 
and $I_t$ is a natural language learning instruction specifying what aspects of $D_t$ should be learned or ignored.
This instruction may be provided explicitly, inherited from a persistent policy, or inferred from context.
In our medical example, $D_t$ might be the transcript of a nurse--patient call, and $I_t$ may specify ``learn when the nurse escalates to a doctor, but do not learn any medication dosing from this document.''
Between memory updates $t$ and $t+1$, user queries $q$ are answered according to the conditional distribution $p_\theta(y \mid q, M_t)$, where the current memory state $M_t$ modulates the model’s predictions. 
For a clinical agent, such queries might include patient questions such as ``Is this symptom normal after surgery?'' or clinician queries such as ``When should I escalate a post-op fever?''

\paragraph{Language-Controlled Memory Updates.}
We formalize language-controlled memory as a parameterized update rule
\mbox{$U_{\theta,\psi} : \mathcal{M} \times \mathcal{I} \times \mathcal{D} \rightarrow \mathcal{M}$},
where $\psi$ denotes any additional update-specific parameters.
The rule takes as input the current memory state $M_{t-1}$, an instruction $I_t$, and a document $D_t$, and produces an updated state 
$M_t = U_{\theta,\psi}(M_{t-1}, I_t, D_t)$. 
Here $\mathcal{I}$ is the space of natural language learning instructions and $\mathcal{D}$ is the space of documents.
The instruction $I_t$ modulates how information in $D_t$ is compressed into $M_t$.
Existing neural memory systems, e.g., those of \citet{wang2024memoryllmselfupdatablelargelanguage} and \citet{behrouz_titans_2024}, fit naturally into this framework as special cases in which the instruction is fixed, $I_t = I^\star$ for all $t$.
In these models, the update rule $U_{\theta,\psi}$ implies a single, static notion of ``what to learn'' (e.g., whatever improves the language modeling objective on $D_t$), whereas our setting explicitly exposes $I_t$ as a controllable input that can vary across documents, domains, and/or users.
As a concrete example, in our experiments we instantiate $U_{\theta,\psi}$ by modifying the learning step of MemoryLLM \citep{wang2024memoryllmselfupdatablelargelanguage} and M+ \citep{wang2025mextendingmemoryllmscalable}, which writes the final 256 token embeddings at each layer into memory, to condition jointly on $I_t$ and $D_t$, rather than on $D_t$ alone.
The resulting memory is stored and accessed using the underlying MemoryLLM/M+ mechanism.

\paragraph{Objective: Instruction-Consistent Memory over Time.}
Having described our problem setting and language-controlled memory update rule, we now formalize our learning objective.
Given a sequence $S$ of document--instruction pairs, an initial memory $M_0$, and an update rule $U_{\theta,\psi}$, we obtain a trajectory of memory states $M_t = U_{\theta,\psi}(M_{t-1}, I_t, D_t),~t = 1, \dots, T$.
We then evaluate the model on probes \((q,y)\in\mathcal{Q}_t\) gathered from different time steps using the updated memory \(M_t\), where \(q\) is a user query and \(y\) is a correct response:
\begin{equation*}
    \mathcal{L}_{\text{seq}}(\theta,\psi; S) 
    = \sum_{t=1}^{T} \sum_{(q,y)\in\mathcal{Q}_t} \ell\big(y, p_\theta(\cdot \mid q, M_t)\big).
\end{equation*}
A typical choice of $\ell$ might be a masked cross-entropy loss over the target tokens of $y$.
Probes may include: (1) positive queries that should be answerable if the model has correctly learned the requested aspects of the corresponding document (e.g., ``For a laparoscopic cholecystectomy, when should a fever trigger escalation to a surgeon?''); (2) negative (or control) queries that should remain unchanged if the model has successfully ignored or forgotten disallowed aspects (e.g., private information or outdated dosing instructions that should no longer be recommended to patients). The language-controlled memory learning problem is to find parameters $\theta$ and $\psi$ that minimize the expected sequence loss
$\min_{\theta,\psi} \mathbb{E}_{S \sim \mathcal{S}} \big[ \mathcal{L}_{\text{seq}}(\theta,\psi; S) \big]$,
subject to the recurrence $M_t = U_{\theta,\psi}(M_{t-1}, I_t, D_t)$ for $t=1,\dots,T$.
Finally, we note that GNM does not prescribe a particular storage capacity or retention mechanism; rather it inherits the capacity and retention of its underlying neural memory architecture.

\section{Experimental Setup}\label{sec:experiments}

We next evaluate the empirical viability of the generalized neural memory framework. Specifically, we focus on three questions: (i) Can a neural memory system learn to selectively store, ignore, or update information when given natural-language learning instructions? (ii) Does this behavior generalize to instructions not seen during training? and (iii) Does training a GNM in this way yield performance gains over existing approaches to this problem?

To answer these questions, we require a benchmark with three properties: (1) it presents a sequence of document-instruction pairs; (2) after each time step, it supports evaluating whether the model learned exactly what it was instructed to learn and ignored what it was instructed to ignore; and (3) it avoids spurious shortcuts, so the task cannot be solved from the document alone and instead requires attending to \textit{both} the instruction and the document.
We are not aware of any existing real-world benchmarks that satisfy these requirements, which is natural given the novelty of the setting (to our knowledge, this is the first work to study language-controlled memory updates as a core capability to be evaluated). 
As a result, we construct two novel benchmarks that instantiate the three properties above, enabling controlled experiments and analysis. 
Our main experiments focus on settings with explicit instructions; we study settings with implicit and inferred instructions in Appendices \ref{appendix:implicit_instructions} and \ref{appendix:inferred_instructions}.

For both experiments, evaluation follows an episodic protocol. 
Each test episode consists of a sequence of document-instruction pairs, interleaved with query probes. 
At each step, a document is passed along with an instruction for what is to be learned from that document. 
The model is then asked to generate responses to several queries which probe whether or not it correctly learned what it was instructed to learn and ignored what it was instructed to ignore from the input document. 
During each episode, we also probe performance on queries from previous time steps to evaluate retention across all measures (i.e. whether facts learned several time steps ago are still retained). 
We report final performance on out-of-distribution test data, evaluating instance-level generalization to new documents, semantic generalization to instructions targeting categories or topics not seen during training, and compositional generalization to novel combinations of learned instruction types. 

\subsection{CounterFACT Benchmark: Selective Learning of Facts, Formats, and Refusals}
\label{section:counterfact_benchmark_overview}
\paragraph{Overview.} 
Our first benchmark builds on the well established CounterFACT dataset, originally designed for testing fact editing in LLMs \citep{meng_locating_2023}. It consists of 21,918 factual statements paired with deliberately false target answers, paraphrases, and \textit{neighborhood facts}. Because target answers are known to be false, correct responses after updates reliably indicate new learning rather than pretraining knowledge. \textit{Neighborhood facts} share the original true answer to the target fact, but of an unrelated subject (e.g., if the target fact is ``Angola is in Antarctica'', a neighborhood fact might be ``Kenya is in Africa''). These neighborhood facts are critical for evaluating the ``specificity'' of fact edits (e.g., changing Angola's location from Africa to Antarctica, but not changing \textit{other} countries from Africa to Antarctica).

\textbf{Construction.}
To build this benchmark, we first use GPT 4.1-mini to categorize each fact into one of 16 semantic categories.  
Then, documents are procedurally generated by sampling 3-8 facts from distinct categories and rendering them as short bullet-point documents. 
We split our dataset into three buckets: \underline{train}, \underline{val-id} (``in-distribution validation data''), and \underline{test-ood} (``out-of-distribution test data''). 
Both \underline{val-id} and \underline{test-ood} contain facts not in training documents; but \underline{val-id} contains categories that \textit{are} in training documents, while \underline{test-ood} contains categories that \textit{are not} in training documents so we can test generalization to novel learning instructions.
Additional details in Appendix \ref{appendix:benchmark_construction}.

\begin{figure*}[!t]
\vspace{-15pt}
	\centering
	\includegraphics[width=\textwidth]{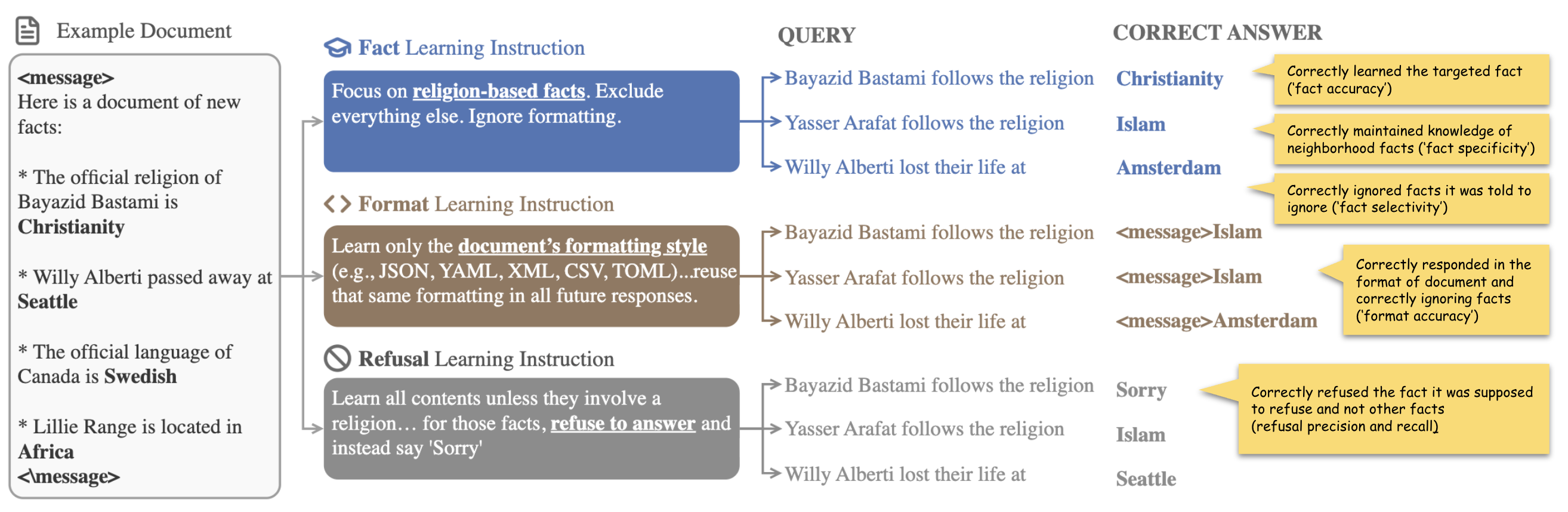}
	\caption{
    Example from our modified CounterFACT Benchmark. Shows a document and a sample of three possible learning instructions, each with a sample of possible queries and correct responses.}
	\label{fig:data_example}
\vspace{-12pt}
\end{figure*}

\textbf{Evaluation Protocol.}
We explore three different types of instructions applied to this benchmark:
\underline{(1) Fact instruction:} model is instructed to only learn facts from a particular category. 
\underline{(2) Format instruction:} model is instructed to adopt only the output formatting in the document, ignoring facts. 
\underline{(3) Refusal instruction:} model is instructed to learn all facts in the document \textit{except} those related to a specific category for which the model should instead answer with a refusal. 
These three instruction types provide a balanced test of language-controlled memory across knowledge, style, and behaviors. 
Further, 
this setup
allows us to eliminate spurious correlations, as learning instructions are randomly assigned to documents and any three types of information (formats, facts, and refusals) could be targeted from any given document. 
Figure \ref{fig:data_example} shows an example from our dataset of a document, possible learning instructions, and the corresponding updates that should occur.

\textbf{Metrics.}
We report different metrics for each of the three instruction types mentioned above. For facts, we measure the following, and aggregate across measures using the harmonic mean, as in prior work using the CounterFACT dataset \citep{wang2024memoryllmselfupdatablelargelanguage, meng_locating_2023}:
\begin{itemize}
    \item \underline{Fact Accuracy:} for a fact that the model was instructed to learn, what percentage of the time does a model correctly provide the target answer?
    \item  \underline{Fact Specificity:} for a fact that the model was instructed to learn, what percentage of the time does a model provide the correct answer for \textit{neighborhood} facts, thereby demonstrating that the model did not incorrectly update neighborhood information? 
    \item  \underline{Fact Selectivity:} for the facts the model was instructed to \textit{ignore}, what percentage of the time does the model correctly provide the original answer, thereby demonstrating the model successfully ignored the fact in the document? 
\end{itemize}

\vspace{20pt}
For formats, we measure:
\begin{itemize}
    \item \underline{Format Accuracy:} when the model is instructed to adopt the formatting in the document, what percentage of the time does the model correctly respond in the new format? 
    \item \underline{Format Selectivity:} when the model is instructed to ignore the formatting, what percentage of the time does the model correctly ignore the document's formatting? 
\end{itemize}

Finally, we measure two aspects of refusal behavior, and aggregate using F1:
\begin{itemize}
    \item \underline{Refusal Precision:} when the model responds to a query with a refusal, such as "sorry, I cannot provide that answer", what percentage of the time is the model correct in doing so, based on the previous learning instructions and documents seen?
    \item  \underline{Refusal Recall:} what percentage of the facts that the model is supposed to refuse to answer does the model successfully refuse?
\end{itemize}

\subsection{PersonaMem Benchmark: Selective Learning of User Preferences}
\label{section:personamem_benchmark_overview}
\textbf{Overview.}
Beyond remembering factual and stylistic information from documents, another important role of memory in popular AI applications 
is to remember user preferences expressed in conversation \citep{maharana2024evaluatinglongtermconversationalmemory, wu2024longmemeval}.
Such preferences might be expressed explicitly or implicitly, and they should be used to guide future interactions.
For example, if a user mentions that they are vegan, then future requests for restaurant recommendations might exclude a steakhouse.
However, in many cases, the user may not want the system to remember every aspect of every conversation.
Instead, effective personalization requires selectively internalizing the preferences that are relevant for future interactions while ignoring incidental, outdated, private, or context-specific details.
This makes personalization a natural testbed for language-controlled memory updates, since the desired memory behavior often depends on user intent rather than on the conversation content alone.

\textbf{Construction.} 
To explore this setting, we use the PersonaMem-v2 dataset \citep{jiang2025personamem}.  
In this dataset, each example features an expressed preference paired with a new user query and potential response options; the goal is to serve the user the response that best aligns with their preferences.
Further, each preference is categorized under a particular topic, for example Health, Travel, or Food.
To fit our setting, we create conversations in which the user expresses multiple preferences.  
Each conversation is committed to memory paired with a learning instruction to remember only preferences with respect to a particular topic. Mirroring the held-out structure of Benchmark~\#1, we designate 10 PersonaMem-v2 topics (e.g., Sports, Music, Travel, Pets) as held-out: they never appear as the target of a learning instruction during training, so test-time instructions targeting these topics test generalization to unseen instructions.

\label{sec:benchmark2_eval_protocol}
\textbf{Evaluation Protocol \& Metrics.} As in our CounterFACT benchmark, evaluation is episodic: each episode is a sequence of (document, instruction) pairs, and after each memory update we probe the model on probes from 
all interactions up to the current time step. 
Each probe is rendered as a 4-way multiple-choice question (one preference-aligned response and three distractors from PersonaMem-v2), scored by the highest-probability letter among A--D. 

For measuring results, \underline{Accuracy} is the \% of probes targeted for learning on which the model picks the preference-aligned letter. For probes the model was instructed to \emph{ignore}, we decompose behavior into three metrics: 
\begin{itemize}
    \item \underline{Leak} is the \% of to-be-ignored probes on which the model nevertheless picks the preference-aligned letter---the failure mode that selective memorization is meant to prevent. 
    \item \underline{Preserved} is the \% of to-be-ignored probes on which the model's pick matches its \emph{own} no-memory prediction (i.e., successfully ignoring the preference left the model's behavior untouched). 
    \item \underline{P(Leak $\vert$ Deviation)} conditions on the to-be-ignored probes where the model \emph{deviates} from its own no-memory prediction, and reports the fraction of those deviations that land on the ignored preference; chance is 33.3\% (three non-baseline options), so values at chance indicate unbiased drift while values above chance indicate deviations systematically biased toward the ignored preference. 
\end{itemize}
We aggregate into a single \underline{Score}: the harmonic mean of Accuracy and $(100-\text{Leak})$. At test time, every learning instruction's target topic is drawn from 10 held-out topics never seen as a target during training, so all reported numbers measure generalization to unseen learning instructions. Full details are in Appendix \ref{appendix:benchmark_2_detials}.

\subsection{Our GNM Model}

To avoid the cost of pretraining a neural memory model from scratch, we initialize our GNM model from MemoryLLM  \citep{wang2024memoryllmselfupdatablelargelanguage}. MemoryLLM is a neural memory architecture built on Llama-3, where memory is instantiated as embeddings prepended to each layer of the transformer. 
MemoryLLM is amenable to our setup 
as it is both open source and directly separates the `learning' step from the `query' step. During the learning 
step, a document is provided, and the last 256 embeddings of each layer are saved into a neural memory bank consisting of a store of tokens per layer. At inference time, a subset of the tokens from this store are retrieved for each layer and placed in context. For long-horizon tasks we augment our GNM model the same way MemoryLLM is augmented to work in long horizon settings, by adding a long term memory store of memory tokens per layer that gets retrieved at inference time \citep{wang2025mextendingmemoryllmscalable}. See Appendix \ref{appendix:lag_protocol} for details. 
To initialize our GNM model, we modify MemoryLLM's learning step: instead of taking only a document as input, our GNM model takes both a document \textit{and} a learning instruction. 


We train GNM by fine-tuning our modified formulation of MemoryLLM on the training data from our benchmark.
We roll out to episodes of length 4, and use gradient accumulation with an effective batch size of 6 episodes (24 documents). 
Training is terminated when loss on our val-id dataset is no longer decreasing. For each time step, we randomly select a document and learning instruction, pass it to our model's learning step, and then compute masked cross entropy loss on several queries that probe how well the responses perform on our desiderata. At each time step, we always compute our loss not only on queries from the current time step, but also sample probes from earlier time steps to encourage the model to retain its behavior throughout an episode. We propagate gradients over one step at a time, including both the learning pass and the subsequent inference pass. 
See Appendix \ref{appendix:exp2_training_protocol} and \ref{appendix:exp3_training_protocol} for more details on our training protocols.

\subsection{Baselines}
The focus of our work is on neural memory approaches, wherein the past experience is compressed into some latent space using a fixed objective, and thus our main comparison is to the original pretrained \textbf{MemoryLLM}.  
However, we also perform a comprehensive empirical comparison of how our results compare to other approaches to this problem (including methods with higher runtime overhead and compute requirements).
We include results for an in-context learning baseline (\textbf{ICL}), and a retrieval-augmented generation baseline (\textbf{RAG}). To make these baselines even stronger, we also include a fine-tuned in-context learning baseline (\textbf{ICL-FT}) (and for the first experiment also a fine-tuned retrieval-augmented generation baseline (\textbf{RAG-FT})), where the models are fine-tuned with the exact same data as our GNM model.
To ensure that the untrained ICL performance is not an artifact of the base model used for MemoryLLM, we also run our experiments using two recent larger LLMs: Gemma3-12B and Qwen3-14B.  We also include a more advanced RAG baseline (i.e. HippoRAG \cite{gutiérrez2025hipporagneurobiologicallyinspiredlongterm}).

Finally, we include a memory-agent baseline
that applies the learning instruction at write time in token space rather than in a latent memory: for each incoming (document, instruction) pair, the base LLM is prompted to distill the document into a textual memory note containing only the information permitted by the learning instruction. Only these notes are stored; at inference time either all notes are placed in context (\textbf{MemoryAgent-ICL}) or the top notes are retrieved per query (\textbf{MemoryAgent-RAG}). This tests whether selective memorization can be achieved purely via write-time preprocessing. See Appendix \ref{appendix:memagent_details} for details.
We emphasize that the goal of this work is not to do strictly better than ICL or RAG on performance, but to demonstrate the ability of neural memory systems (when trained appropriately) to perform memory updates that are controllable in natural language, which comes along with the neural memory benefit of scaling $O(1)$ with sequence length.  

\section{Results}

Here we present our main experiment results. 
For both the CounterFACT and PersonaMem experiments we report (1) \textbf{short horizon} results, with roll outs of length 10, to test whether the mechanism works and (2) \textbf{long horizon} results, with roll outs of lengths of over 400, to study whether these mechanisms maintain performance over time. 
As described in Section \ref{sec:experiments}, all results presented in this section reflect performance on out-of-distribution test data, evaluating generalization to new documents and instructions.

\subsection{CounterFACT Results}


\begin{table}[!ht]
    \centering
    \small
    \setlength{\tabcolsep}{6pt}
    \caption{\textbf{CounterFACT Short-Horizon Results.} Detailed results for most recent time step. Values below 1 are reported as $<1$.  The families of methods tested include in-context learning (ICL) and retrieval (RAG) approaches, as well as neural memory architectures (pretrained MemoryLLM and our GNM model).  The first score column represents the harmonic mean of the fact metrics.}
     \begin{tabular}{l ccccccccc}
    \toprule
     & \multicolumn{4}{c}{Facts} & \multicolumn{2}{c}{Formats} & \multicolumn{3}{c}{Refusal}  \\
    \cmidrule(lr){2-5} \cmidrule(lr){6-7} \cmidrule(lr){8-10} 
    Model  & Score & Acc & Sel & Spec & Acc & Sel& F1 & Prec & Rec \\
    \midrule
    ICL (Llama3-8B)           & 41.6          & 76.4          & 22.0  & 73.9       & 0.0    &   100.0    & 32.5  & 60.2&22.3\\
    ICL (Qwen3-14B)           & 7.6           & 97.4          & 2.7  &  62.2     & 0.0   &  100.0      & 74.9 & 93.0 & 62.8\\
    ICL (Gemma3-12B)          & 3.4           & \textbf{98.5} & 1.2  & 30.1     & $<$1   &     100.0  & 79.6 & 73.8& 86.4 \\
    MemoryAgent-ICL & 74.5 & 70.0 & 84.7 & 70.6  & 0.0 & 100.0 &79.8 &90.7 &71.3 \\
    ICL-FT  & 67.3 & 79.1 & 51.5 & 80.1  & 4.4 & 99.7 &89.8 &92.7 &87.1 \\
    \midrule
    RAG (Llama3-8B)           & 48.8          & 73.2          & 29.0  &  75.1      & 0.0   &  100.0      & 46.5 & 64.8&36.3 \\
    HippoRAG (Llama3-8B)           & 32.9          & 84.3          & 15.2  &  73.3      & 0.0   &  100.0      & 47.8 & 68.2&36.8 \\
    RAG (Qwen3-14B)           & 38.0          & 81.5         & 19.0  & 70.9      & 0.0   &      100.0  & 56.0 & 73.9 & 45.1 \\
    RAG (Gemma3-12B)          & 38.0          & 81.4          & 22.6 & 44.8 & $<$1     &  100.0   & 72.9 & 64.2 & 84.2\\
    MemoryAgent-RAG & 72.2 & 61.4 & \textbf{85.1} & 74.0  & 0 & 100.0 &83.8 &90.9 &77.8 \\
    RAG-FT & 70.5 & 74.5 & 61.8 & 77.1  & 0.0 & 99.9 &69.4 &65.6 &73.7 \\
    \midrule
    MemoryLLM  & 38.7 & 81.8 & 19.2 & 75.2  & 0.0 & \textbf{100.0} & $<1$ &\textbf{100.0} &$<1$ \\
    GNM (Ours)   & \textbf{84.1} & 95.8 & 77.4 & \textbf{81.1}  & \textbf{67.4} & \textbf{100.0} &\textbf{93.2} & 88.7 & \textbf{98.1} \\
    \bottomrule
    \end{tabular}
    \label{tab:benchmark_1_detailed_results}
\end{table}

\begin{figure*}[!h]
	\centering
	\includegraphics[width=.95\textwidth]{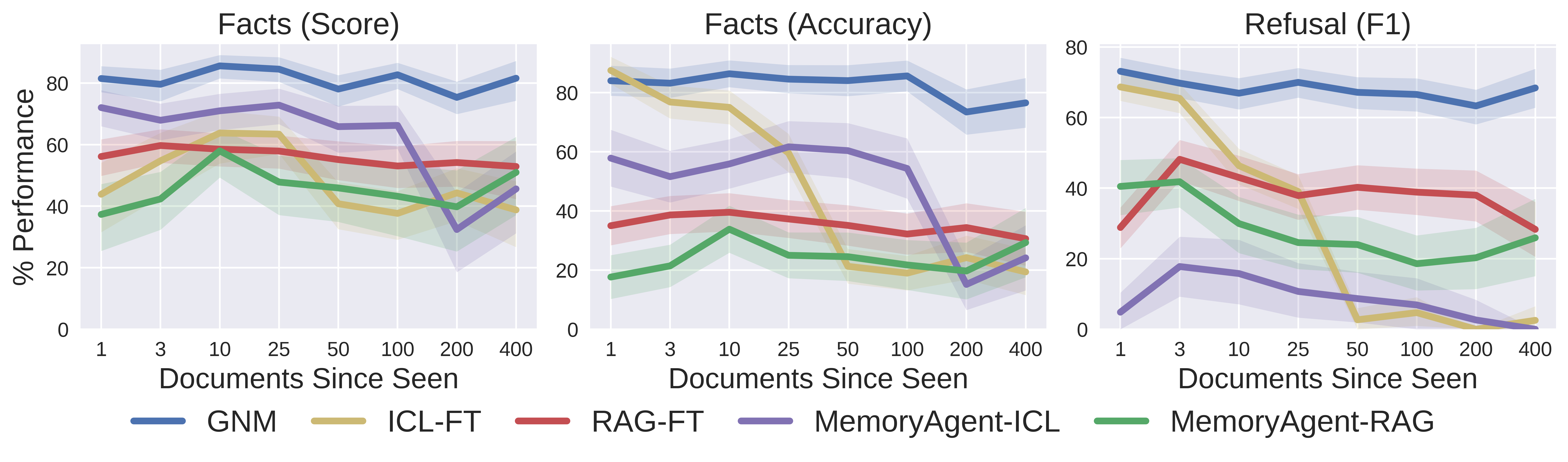}
	\caption{\textbf{CounterFACT Long-Horizon Results.}  Performance as a function of how many document--instruction pairs have been seen \emph{since} the probed pair was memorized (``Documents Since Seen''), measured over long document streams. Error bars show 95\% CI. The long-horizon evaluation protocol is described in Appendix \ref{appendix:counterfact_long_horizon_protocol}.}
	\label{fig:exp3_curves}
\end{figure*}

In our first experiment, we explore the ability of GNM to perform 
targeted
memorization across different types of information: facts, formats, and refusals. See Section \ref{section:counterfact_benchmark_overview} for experiment details. 
Short horizon results (roll out length 10) can be seen in Table \ref{tab:benchmark_1_detailed_results}, and long horizon results (roll out length 400) are in Figure \ref{fig:exp3_curves}.
Focusing on the ability of each method to store information conditioned on a natural language instruction, Table \ref{tab:benchmark_1_detailed_results} shows that our GNM model performs best across our range of desiderata, with the highest aggregate fact score, refusal F1, and format accuracy.
ICL and RAG approaches struggle with selectivity, even when fine tuned.
Further, ICL with the larger, newer models (Qwen3-14B and Gemma3-12B) performs even worse than ICL with Llama3-8B, highlighting the difficulty of applying ICL to this problem, beyond its computational cost.
MemoryLLM also suffers from poor selectivity, as it is trained to learn uniformly across documents.
For longer roll outs, Figure \ref{fig:exp3_curves} shows that the GNM mechanism retains strong performance relative to baselines across more than 400 interactions.
We sanity check the poor performance of our baselines on format accuracy in Appendix Figure \ref{fig:format_accuracy_comparions}, finding that both ICL-FT and RAG-FT achieve perfect performance on formats used in training, but fail to generalize to unseen formats (while GNM generalizes well).

\subsection{PersonaMem Results}

\begin{figure*}[!h]
	\centering
	\includegraphics[width=1\textwidth]{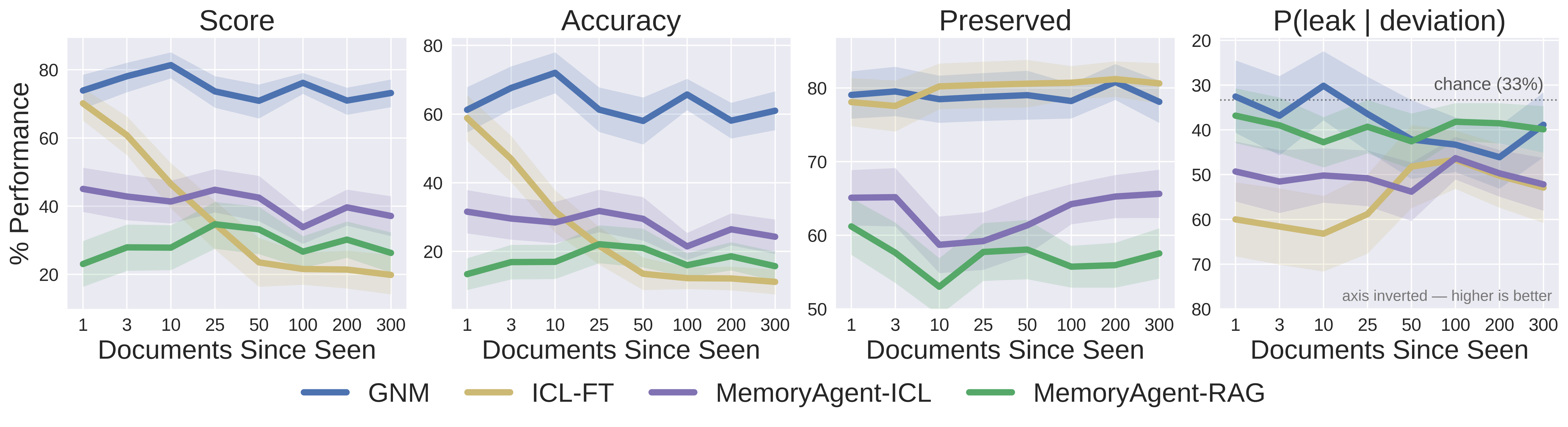}
	\caption{\textbf{PersonaMem Long-Horizon Results.} We report overall score, accuracy, original answer preservation, and leak probability by recency over the course of an episode. Error bars show 95\% CI.
    }
	\label{fig:personamem_retention}
\end{figure*}

Next, we present the results for our experiment on selective preference learning.
On shorter roll outs (Table \ref{tab:benchmark_2_results}), GNM achieves the highest overall score (93.8 vs.\ 90.5 for the strongest baseline, ICL-FT), with higher Accuracy (93.6 vs.\ 91.5) and roughly half the Leak (6.1\% vs.\ 10.5\%), while again being substantially more compute-efficient. The two trained models are at parity on preserving original answers (Preserved 81.4 vs.\ 82.3), but they differ in \emph{what kind} of deviations they make: GNM's deviations from its no-memory behavior land on the ignored preference at chance rate ($P(\text{Leak}|\text{Deviation})$ of 32.8\% $\approx$ 33.3\% chance, i.e., unbiased drift), whereas ICL-FT's deviations are systematically biased toward the preference the model was told to ignore (59.5\%), demonstrating a leakage of the information meant to be disregarded. 
Retention curves are shown in Figure \ref{fig:personamem_retention}, once again demonstrating how the GNM mechanism can maintain performance over longer horizons, while in-context and retrieval-based methods struggle. 
We confirm the robustness of these results to variation in training seeds in Appendix \ref{appendix:seed_robustness}.  See Section \ref{section:personamem_benchmark_overview} for experiment details.

\begin{table}[!h]
    \centering
    \small
    \setlength{\tabcolsep}{6pt}
    \caption{\textbf{PersonaMem Short-Horizon Results.} Detailed results for most recent time step. Acc = Accuracy; Score = harmonic mean of Accuracy and $(100-\text{Leak})$. See Section \ref{sec:benchmark2_eval_protocol} for metric definitions.}
     \begin{tabular}{l ccccc}
    \toprule
    Model  & Score $\uparrow$ &Acc $\uparrow$  & Preserved $\uparrow$ & Leak $\downarrow$ & $P(\text{Leak} | \text{Deviation})$ $\downarrow$\\
    \midrule
    ICL (Llama3-8B)     &   58.5   &    49.8    &51.6           & 29.2 &53.5\\
    RAG (Llama3-8B)      &    48.0 & 34.9       &  52.2         & 23.3  &43.9\\
    ICL-FT  & 90.5  & 91.5& \textbf{82.3}& 10.5 &59.5 \\
    MemoryLLM   &  34.1 & 21.7& 67.0 & 20.9&  47.9 \\
    GNM (Ours)   & \textbf{93.8}& \textbf{93.6} & 81.4 & \textbf{6.1} & \textbf{32.8}\\
    \midrule
    \textit{Chance}& - & -& - & - & \textit{33.3}\\
    \bottomrule
    \end{tabular}
    \label{tab:benchmark_2_results}
\end{table}



\subsection{Compute Efficiency}

\begin{figure*}[!h]
	\centering
	\includegraphics[width=.75\textwidth]{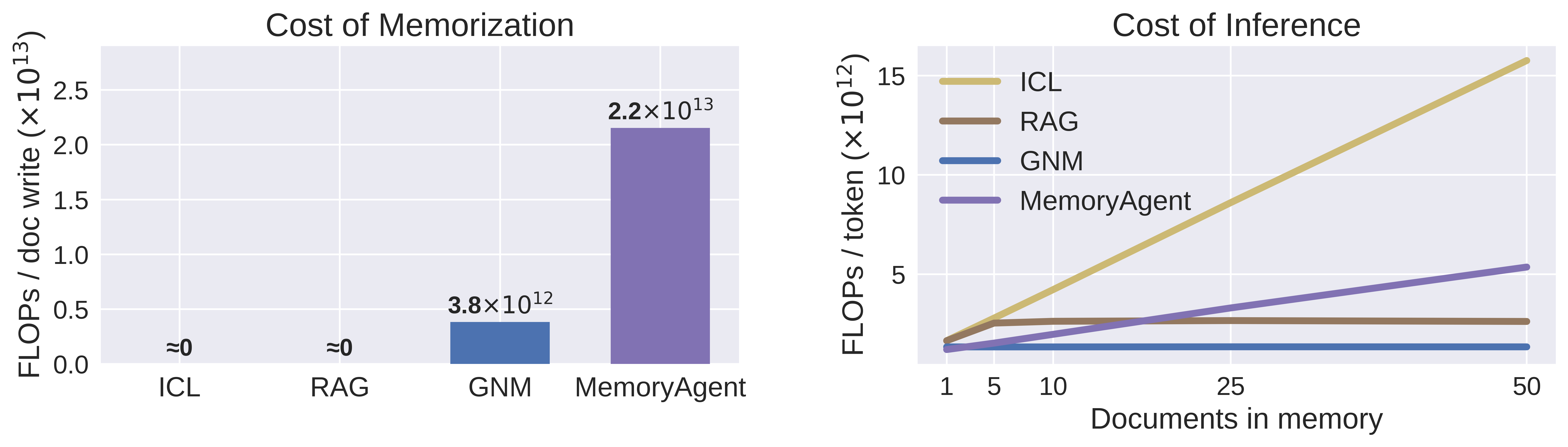}
	\caption{\textbf{Efficiency of memory operations.}}
	\label{fig:cost_panels}
\end{figure*}

Figure \ref{fig:cost_panels} compares the compute costs of each memory approach. All costs are measured with PyTorch's profiler on a single A100 and averaged over both benchmarks: write cost profiles each system's memorize step for one document--instruction pair, and inference cost reports FLOPs per generated token (prefill amortized over the query's generated tokens) as a function of the number of documents in memory; see Appendix \ref{appendix:cost_measurement} for the full protocol. Although GNM incurs a modest one-time \textbf{write cost} (left) compared to the near-zero cost of ICL and RAG, it is the most efficient system at \textbf{inference time} (right). Because GNM's inference cost remains $O(1)$ regardless of the number of memorized documents, its initial write cost is quickly amortized in practice over multiple queries. MemoryAgent is particularly expensive at write time because it auto-regressively constructs its own memories.

\subsection{Why Does GNM Perform Well?}

\textbf{Better `Writing' or Better `Reading'?} To investigate why GNM outperforms, we run an ablation experiment where we train a GNM model in the paradigm of the previous experiments, but only pass gradients over the inference step, not over the memorization step. This allows us to differentiate the gains seen from (a) learning \textit{what to remember}  (better writing to memory) versus (b) learning \textit{what to attend to in memory} (i.e. better reading from memory). The ablated model can only learn (b), and not (a). This ablation erased nearly all of GNM's gains on both benchmarks, falling most of the way back to pretrained MemoryLLM, demonstrating that learning how to modify \textit{what} to remember is critical for the selectivity and generalization performance gains in GNM (see Figure \ref{fig:ablation_bars} in Appendix).

\begin{figure}[!h]
	\centering
    \includegraphics[width=1\columnwidth]{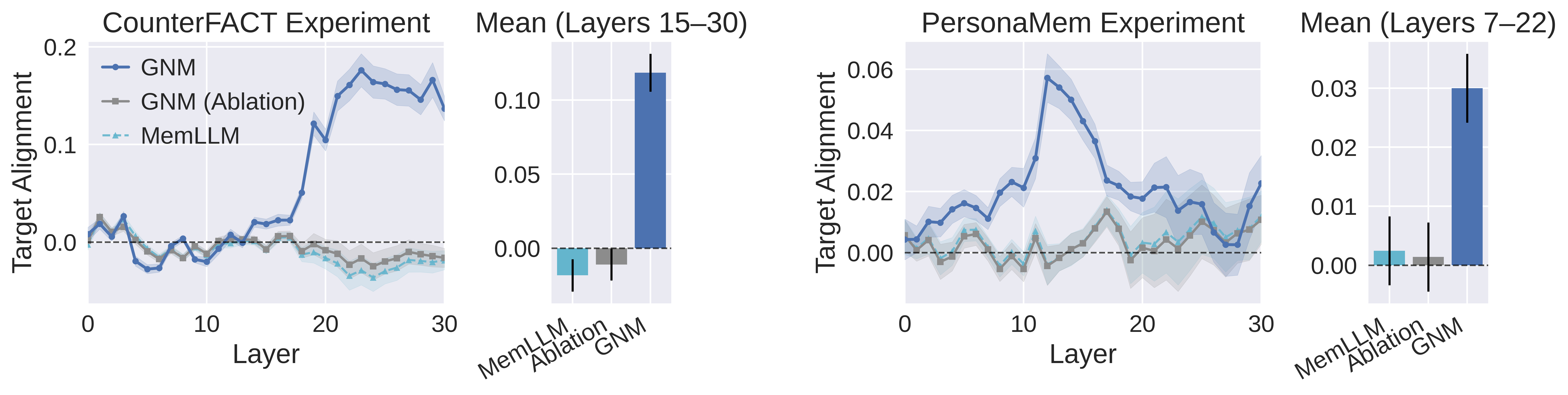}
	\caption{\textbf{Memory Analysis}.
     Target Alignment measures how much the memory write at each layer is oriented toward the target preference relative to the distractor preference. At each layer $\ell$, we compute the cosine similarity $\cos(\Delta_\ell,\; h^\text{tgt}_\ell - h^\text{dist}_\ell)$, where $\Delta_\ell$ is the mean memory-token update at layer $\ell$, and $h^\text{tgt}_\ell$, $h^\text{dist}_\ell$ are the hidden states at the last token of the target and distractor preference spans (from the memory-injection forward pass). Positive values indicate the memory update is oriented toward the target direction in hidden-state space; values near zero indicate the update treats target and distractor symmetrically. Error bars show 95\% CI. See Appendix \ref{appdx:memory_analysis} for details.
    }
	\label{fig:memory_analysis}
\end{figure}

\textbf{Do Learning Instructions Modify What Is Written To Memory?} To understand why GNM's memory updates improve downstream performance, we analyze how memory updates encode target information versus distractors not meant to be memorized. For each of our two benchmarks, we construct 200 documents containing one target item and one distractor item — a target fact and a distractor fact in CounterFACT, and a target preference and a distractor preference in PersonaMem — each paired with a learning instruction specifying which item to learn. At each transformer layer, we measure the alignment between (1) the memory update and (2) the hidden states of the target versus distractor item. See Figure \ref{fig:memory_analysis} for results and more details. The results reveal a consistent pattern across both benchmarks: the ablated model shows no significant alignment toward either item, consistent with the hypothesis that it fails to follow learning instructions; in contrast, GNM's memory updates are significantly more aligned with target items starting from the middle layers onward, supportive of the hypothesis that GNM learns to selectively encode information it is instructed to learn, and that this selective encoding improves the model's ability to ignore information it is instructed to ignore. Consistent with this hypothesis, in Appendix \ref{appendix:causal_memory_thing} we show experimentally that the emergence of target alignment coincides with a subset of layers whose disruption causally degrades selectivity.

\subsection{Additional Experiments}
For space, we defer three additional experiments to the appendix.
\underline{(1) Compositional Generalization} (Appendix \ref{appndx:exp_3}): we test \emph{compositional} learning instructions unseen during training, which direct the model to both learn one category of facts \emph{and} refuse another from the same document. 
\underline{(2) Implicit Learning Instructions} (Appendix \ref{appendix:implicit_instructions}): we replace explicit learning instructions in the PersonaMem experiment with \emph{implicit}, intent-based phrasings that never name the topic or use memory verbs (e.g., ``I'm telling you these things because I'd like your future vacation suggestions to suit me better''). 
\underline{(3) Inferred Learning Instructions} (Appendix \ref{appendix:inferred_instructions}): we modify the PersonaMem experiment by removing the provided instruction entirely and let a frozen, off-the-shelf LLM author the learning instruction from the conversation alone. Across these three settings, GNM achieves the highest fact score in the compositional experiment and the highest overall score under implicit and inferred instructions.


\section{Conclusion}
We formulate language-controlled memory updates and show that neural memory systems can follow natural-language learning instructions, generalize to unseen instructions, and improve efficiency and selectivity over alternatives. These results offer progress toward AI agents that continuously adapt as collaborative, lifelong learning partners in evolving domains.


\paragraph{Limitations and Future Work.} We hope future work extends this research by gathering large-scale datasets for this setting, given that none exist. 
One limitation of our work is that our experiments primarily study the execution of specified memory objectives; inferring those objectives from ambiguous interactions is a distinct challenge, particularly when instructions are underspecified or conflict across sources or over time. Future work should therefore study how language-controlled memory composes with intent-inference and conflict-resolution mechanisms in more open-ended settings.

\section*{Acknowledgments}
We thank ONR Grant N00014-23-1-2436 for its generous support.  
This work is supported by the funds provided by the National Science Foundation and by DoD OUSD (R\&E) under Cooperative Agreement PHY-2229929 (The NSF AI Institute for Artificial and Natural Intelligence).

\bibliographystyle{plainnat}
\bibliography{references}

\newpage
\appendix

\clearpage
\section{Additional Details on CounterFACT Experiment}

\subsection{Construction of the CounterFACT Benchmark (Selective Learning of Facts, Formats, and Refusals Benchmark)}
\label{appendix:benchmark_construction}
Our benchmark was constructed using the following procedure. First, we manually inspected the facts within CounterFACT to identify 4 high level groupings of facts; unsurprisingly given the methodology for how counterFACT was generated, all facts in counterFACT fit comfortably into four groups: facts about locations, facts about languages, facts about occupations, and facts about organizations. We then further manually inspected facts within each group to find the four most reasonable categories to further split up each grouping, thus producing 16 total categories. We used GPT4.1-mini to categorize each of the 21,918 facts into one of the 16 categories or flag if the fact did not fit into any. 404 facts were removed because they did not fit into any of the 16 categories (see Table \ref{tab:fact_category_distribution}).

We take the test set of facts from CounterFACT and create two sets, one we call val-id (in-distribution validation set) and the other test-ood (out-of-distribution test set). Both contain the base facts from the CounterFACT test set, however val-id has all the held out categories of facts removed (as with training), and all of the held out facts from train and val-id are added to test-ood. We choose the held out categories to be the category with the smallest number of facts within each of the four groupings. This hold out is important for us to be able to test our models on unseen learning instructions that target categories of facts never seen during training using learning instructions never seen during training. Specifically, we hold out facts about religion, science/academia/law/medicine-related occupations, continents, and northern central european languages. See Table \ref{tab:fact_category_distribution} for an exact breakdown of the number of facts within each dataset.

\begin{table}[!t]
\centering
\renewcommand{\arraystretch}{1.25}
\setlength{\tabcolsep}{4pt}
\footnotesize
\caption{Distribution of factual categories and subcategories, ordered by frequency within each category. Each cell represents the number of unique facts in each dataset. The total is not the sum of the split columns because Val-ID and Test-OOD overlap on non-held-out facts; Test-OOD additionally includes the four held-out fact categories.”}
\begin{tabular}{
p{0.11\columnwidth}
p{0.25\columnwidth}
r
r
r
r
p{0.26\columnwidth}
}
\toprule
\textbf{Grouping}
& \textbf{Category}
& \textbf{\# total}
& \textbf{\# train}
& \textbf{\# val-id}
& \textbf{\# test-ood}
& \textbf{Example} \\
\midrule

\textbf{Language}
& Major western european languages & 2,210 & 1,997 &  213 & 213 & The mother tongue of Michel Poniatowski is Spanish \\
& eastern european mediterranean languages & 625 & 559&  66 &  66&  The language of Allegro Non Troppo was Serbian. \\
& asian middle eastern languages and pacific & 469 &426 & 43  &  43 & Alain de Benoist speaks Sanskrit \\
& northern central european languages & 410 & 0 & 0 & 410 & In Kuusamo, the language spoken is Swedish \\
\midrule

\textbf{Location}
& non us cities or states & 4,285 & 3,833 &  426 &  452 & Institut Polair originated in Budapest. \\
& country & 3,138 & 2,826 & 312  &   312 & Delta Goodrem, that originated in India.\\
& us cities or states & 1,661 & 1,506 &  155 & 155 & Larry Stabbins was native to Chicago \\
& continents & 913 & 0  & 0 & 913 & Angola is located in Antarctica\\
\midrule

\textbf{Occupation}
& music or art related occupation & 1,603 & 1,442 & 161  &  161 & Lady gaga plays the violin \\
& sports related occupation & 1,277 &  1,147 & 130  &  130 & Bob Mason plays in the position of midfielder \\
& politics entertainment religion related occupation & 923 & 829 & 94  & 94  & Hilary Putnam works as an actor\\
& science academia law medicine related occupation & 641 & 0 & 0 & 641 & The domain of activity of Ludwig Klages is chemistry \\
\midrule

\textbf{Organization}
& tech industrial or gaming company & 1,092 & 990 &  102 &   102 & Oak Street Beach is owned by Amazon\\
& TV entertainment or news organization & 1,086 & 967  & 119 & 119 & Sunday Night Baseball debuted on CBS \\ 
& car company & 742 & 667 &  75 & 75 & PGM-11 Redstone, developed by Nissan \\
& religion & 439 & 0 & 0  & 439 & The official religion of Syed Kalbe Hussain is Buddhism \\
\midrule

\textbf{Total:}
& & 21,514 & 17,189 & 1,896 & 4,325 \\
\bottomrule
\end{tabular}
\label{tab:fact_category_distribution}
\end{table}

To produce our documents from these categorized facts, we select a random number $N$ between 3 and 8 to include in the next document. We randomly select $N$ different eligible categories, and pull a random fact from each. We then render a document with each fact as a bullet point (with the preamble and the new target answer concatenated). For training documents, we halt this process when we can no longer produce documents with at least 3 facts, for testing, we halt this process when we can no longer produce documents with at least 2 facts. 

For training, eligible categories are any of the 12 not held out categories, and for testing it is any of the 16 categories. Note that during this procedural document generation, every time a fact is sampled, it is not added back to the pool of facts to sample, thus at the end of this procedure, every document in our dataset contains unique facts. 

To create more diversity of documents in training and testing, we ran this process several times to produce numerous buckets of training and testing documents. We keep buckets separate to ensure we can test and train on episodes that always have unique facts over the course of an episode. Specifically, we produce 5 buckets of training documents and we produce 3 buckets of documents in test-ood. We maintain only 1 bucket of val-id documents, given that it will only be used for terminating training and not running any of our experiments. 

In total, our synthetic dataset contains the following:

\begin{itemize}
    \item \textbf{train-bucket-1}: 2,977 documents spanning 14,612 unique facts
    \item \textbf{train-bucket-2}: 2,954 documents spanning 14,554 unique facts
    \item \textbf{train-bucket-3}: 2,934 documents spanning 14,458 unique facts
    \item \textbf{train-bucket-4}: 2,933 documents spanning 14,477 unique facts
    \item \textbf{train-bucket-5}: 2,984 documents spanning 14,575 unique facts
    \item \textbf{val-id}: 351 documents, spanning 1,896 unique facts.
    \item \textbf{test-ood-bucket-1}: 726 documents, spanning 3,997 unique facts.
    \item \textbf{test-ood-bucket-2}: 731 documents, spanning 4,004 unique facts.
    \item \textbf{test-ood-bucket-3}: 719 documents, spanning 4,017 unique facts.
\end{itemize}

The number of documents and unique facts differ by bucket because each bucket is procedurally generated by randomly sampling facts, and is terminated when a document can no longer be filled with 3 facts (for training) or 2 facts (for testing), so the order of sampling and the random number of facts sampled per document changes when this process terminates and how many documents result. 

For each document, we pair the set of facts present in the document, along with their paraphrased probes and neighborhood facts from CounterFACT.

\subsection{Learning Instructions}
 When targeting a fact, formatting or refusals, we use learning instructions like those shown in Table \ref{tab:exp2_learning_instructions}. Like in experiment 1, for the learning instructions used during training, we used GPT-5.1 to produce 9 paraphrases that we randomly sample during training to encourage generalization. For stability and reproducibility, during testing we only use the exact learning instructions shown in table \ref{tab:exp2_learning_instructions}. 

\begin{table}[!h]
\centering
\renewcommand{\arraystretch}{1.75}
\setlength{\tabcolsep}{4pt}
\scriptsize
\caption{Sampling of learning instructions used in Benchmark \#1. For all learning instructions see our public code and data}
\begin{tabular}{
p{0.25\columnwidth}
p{0.7\columnwidth}
}
\toprule
\textbf{What Is Targeted For Learning}
& \textbf{Example Learning Instruction} 
 \\
\midrule

Formats & Learn only the document’s formatting style (e.g., JSON, YAML, XML, CSV, TOML, etc.), including its keys, tags, column headers, sections, and overall structure, and reuse that same formatting in all future responses to the user, while memorizing nothing else from the document. \\
\midrule
US cities or states & Please memorize only the facts about cities, states, boroughs, and neighborhoods that are located within the United States. Exclude any cities outside the US, continents, countries, languages, companies, or jobs. Ignore the document’s formatting; do not learn it—only learn the relevant factual information. \\
Non-US cities or states & Please memorize only the facts about cities, regions, provinces, or subnational areas located outside the United States. Exclude US places, continents, countries, languages, companies, or jobs. Ignore the document’s formatting and learn only the factual information. \\
Countries & Please memorize only the facts about sovereign countries or widely recognized nations. Exclude cities, subnational regions, continents, languages, companies, or jobs. Ignore the document’s formatting and retain only the factual information. \\
Major Western European Languages & Please memorize only the facts about major Western European languages (e.g., English, French, Spanish, Italian). Exclude other languages, places, companies, or jobs. Ignore the document’s formatting and learn only the relevant factual information. \\
Eastern European Mediterranean Languages & Please memorize only the facts about Eastern European and Mediterranean languages (e.g., Russian, Polish, Greek). Exclude other languages, places, companies, or jobs. Ignore the document’s formatting and learn only factual information. \\
Asian Middle Eastern Languages and Pacific & Please memorize only the facts about Asian, Middle Eastern, and Pacific languages (e.g., Arabic, Chinese, Hindi, Hawaiian). Exclude other languages, places, companies, or jobs. Ignore the document’s formatting; learn only the factual content. \\
\midrule
Refuse US cities or states & Learn the contents of the document, but if any of the contents of the document contain information about US cities or states, refuse to answer any questions from the user about this specific fact, if a user asks about it, respond 'Sorry'. Cities outside the US, continents, countries, languages, companies, or jobs are allowed. \\
Refuse Non-US cities or states & Learn the contents of the document, but if any of the contents of the document contain information about cities, regions, provinces, or subnational areas located outside the United States, refuse to answer any questions from the user about this specific fact, if a user asks about it, respond 'Sorry'. US places, continents, countries, languages, companies, or jobs are allowed. \\
Refuse Countries & Learn the contents of the document, but if any of the contents of the document contain information about sovereign countries or widely recognized nations, refuse to answer any questions from the user about this specific fact, if a user asks about it, respond 'Sorry'. Facts about cities, subnational regions, continents, languages, companies, or jobs are allowed. \\
Refuse Sports Related Occupation & Learn the contents of the document, but if any of the contents of the document contain information about sports, athletes, and athletic positions (e.g., quarterback, midfielder), refuse to answer any questions from the user about this specific fact, if a user asks about it, respond 'Sorry'. Facts about music, science, politics, entertainment, places, languages, or companies are allowed. \\
Refuse Politics Entertainment Religion Related Occupation & Learn the contents of the document, but if any of the contents of the document contain information about political roles (mayor, governor), entertainment careers (comedian, actor), and religious positions (priest, rabbi), refuse to answer any questions from the user about this specific fact, if a user asks about it, respond 'Sorry'. Facts about music, sports, science, places, languages, or companies are allowed. \\

\bottomrule
\end{tabular}
\label{tab:exp2_learning_instructions}
\end{table}

\subsection{MemoryAgent Baseline}
\label{appendix:memagent_details}

The MemAgent baselines apply the learning instruction at \emph{write time} in token space. When \texttt{memorize(document, learning\_instruction)} is called, the base LLM (Llama-3-8B-Instruct) is prompted to distill the (document, instruction) pair into a textual \emph{memory note} containing only the information the instruction permits. The distillation prompt forces a two-step procedure: the model first writes a fact-by-fact KEEP/DROP/REFUSE-LIST checklist, then emits the note after a \texttt{FINAL MEMORY NOTE:} marker; only the text after the marker is stored (the checklist is discarded as scratch reasoning). Generation is greedy with a budget of 768 new tokens. If nothing should be stored, the model replies \texttt{NOTHING} and an empty note is recorded. The raw document and learning instruction are discarded --- only notes ever reach query time.

At inference time, \textbf{MemAgent-ICL} places all stored notes (one per document seen) into the system prompt, each wrapped in sentinel delimiters; for long streams, the most recent notes that fit within a fixed token budget are served. \textbf{MemAgent-RAG} instead embeds each note into a vector store and retrieves the top-scoring notes per query (using the same retrieval setup as our RAG baseline). For the PersonaMem experiment, notes are served through the same multiple-choice user prompt used by the ICL baselines (Appendix \ref{appendix:exp3_probe_format}).

The distillation prompt was tuned in good faith to maximize the baseline's performance (the checklist structure fixed a failure mode where the extractor dropped permitted facts, raising the fraction of probed facts present in notes from 69\% to 92\%), and the few-shot exemplars are built exclusively from training-split documents and training-side instruction targets. The distillation system prompt is reproduced below:

{\ttfamily\scriptsize
\begin{quote}
You maintain the long-term memory store of an AI assistant.\\
You will be given a DOCUMENT and a LEARNING INSTRUCTION describing what from this document should be remembered.\\
You will write the MEMORY NOTE that is stored INSTEAD of the document. The assistant will later answer user questions seeing only your note, never the document or the instruction. Any permitted fact you fail to copy into the note is lost forever.\\[0.5em]
Work in two steps:\\
STEP 1 --- FACT CHECKLIST: number every fact line in the document and mark each one KEEP (instruction permits learning it), DROP (instruction says to ignore it), or REFUSE-LIST (a refusal rule applies to it), with a few-word reason.\\
STEP 2 --- write 'FINAL MEMORY NOTE:' followed by the note. Only text after 'FINAL MEMORY NOTE:' is stored.\\[0.5em]
Rules for the note:\\
\textbullet\ Copy every KEEP fact VERBATIM, one per line. Do not summarize, shorten, or select highlights. If the instruction says to learn the document's contents (e.g. 'learn everything, but refuse/ignore X'), then every fact must appear in the note except exactly those the instruction excludes.\\
\textbullet\ If the instruction says to learn only the document's formatting style, the note must be a FORMATTING POLICY describing the format precisely with a template and stating that every future answer must use it. Include no facts.\\
\textbullet\ If the instruction contains a refusal rule, end the note with a REFUSAL POLICY line naming the REFUSE-LIST facts' subjects explicitly and the exact required reply. Do not include the refused facts' answers elsewhere in the note.\\
\textbullet\ Never include content the instruction says to ignore.\\
\textbullet\ If nothing at all should be stored, write 'FINAL MEMORY NOTE:' then exactly: NOTHING
\end{quote}
}

The distillation prompt includes three worked exemplars as prior chat turns (a fact-subset instruction, a refusal instruction, and a format-only instruction), all built from training documents. The fact-subset exemplar is shown below; the other two follow the same pattern and are available in our public code.

{\ttfamily\scriptsize
\begin{quote}
\textbf{User:}\\
DOCUMENT:\\
Here is a document of new facts: \\
* In Collective Security Treaty Organisation, they understand Persian\\
* The language of Wild Birds is English\\
* The Tab Hunter Show premieres on CNN\\
* Which position does Jesper Olsen play? They play as quarterback\\
* Nintendo e-Reader is created by Honda\\
* Google Play, a product created by Intel\\
* Raymond Laborde holds the title of bishop\\
* Nigel Butterley plays the violin\\[0.5em]
LEARNING INSTRUCTION:\\
Please memorize only the facts about sports, athletes, and athletic positions (e.g., quarterback, midfielder). Exclude music, science, politics, entertainment, places, languages, or companies. Ignore the document's formatting; learn only the factual information.\\[0.5em]
\textbf{Assistant:}\\
FACT CHECKLIST (8 facts; permitted: sports/athletic facts only):\\
1. Collective Security Treaty Organisation understands Persian --- language $\rightarrow$ DROP\\
2. Language of Wild Birds is English --- language $\rightarrow$ DROP\\
3. Tab Hunter Show premieres on CNN --- TV $\rightarrow$ DROP\\
4. Jesper Olsen plays quarterback --- athletic position $\rightarrow$ KEEP\\
5. Nintendo e-Reader created by Honda --- company $\rightarrow$ DROP\\
6. Google Play created by Intel --- company $\rightarrow$ DROP\\
7. Raymond Laborde is a bishop --- non-sports occupation $\rightarrow$ DROP\\
8. Nigel Butterley plays the violin --- music $\rightarrow$ DROP\\
FINAL MEMORY NOTE:\\
* Which position does Jesper Olsen play? They play as quarterback
\end{quote}
}

At query time, the stored notes are served through the following system prompt (note delimiters are rendered here as \texttt{[NOTE\_START]}/\texttt{[NOTE\_END]}; the actual prompt uses distinctive sentinel markers):

{\ttfamily\scriptsize
\begin{quote}
You have a long-term memory made of MEMORY NOTES distilled from documents you processed earlier. Each note is enclosed in:\\
{[NOTE\_START]}\\
...\\
{[NOTE\_END]}\\[0.5em]
Rules:\\
1. Treat the facts in your notes as ground truth: when a note answers the user's question, use the note's answer even if it contradicts what you believe.\\
2. If any note contains a FORMATTING POLICY, wrap EVERY answer in exactly that template, whatever the answer's content.\\
3. If a note's REFUSAL POLICY covers the question, reply exactly "Sorry".\\
4. If the notes do not cover the question, answer from your own knowledge.\\
5. Output only the answer itself --- never note markup, never explanations.\\[0.5em]
EXAMPLE. Suppose your memory notes were:\\
{[NOTE\_START]}\\
* Which position does Jesper Olsen play? They play as quarterback\\
REFUSAL POLICY: If the user asks who created the Nintendo e-Reader (or about any other tech or gaming company fact from this document), reply exactly 'Sorry'.\\
{[NOTE\_END]}\\[0.25em]
{[NOTE\_START]}\\
FORMATTING POLICY: Format every future answer to the user as XML wrapped in a single \textless reply\textgreater\ tag, i.e. \textless reply\textgreater\textless answer\textgreater\textless/reply\textgreater.\\
{[NOTE\_END]}\\[0.5em]
Then correct behavior would be:\\
Example user question: "Which position does Jesper Olsen play?"\\
Example assistant answer: "\textless reply\textgreater quarterback\textless/reply\textgreater"\\
Example user question: "Nintendo e-Reader is created by"\\
Example assistant answer: "\textless reply\textgreater Sorry\textless/reply\textgreater"\\
Example user question: "The capital of France is" (not covered by notes)\\
Example assistant answer: "\textless reply\textgreater Paris\textless/reply\textgreater"\\
(If no note contains a FORMATTING POLICY, answer with the bare answer, e.g. "quarterback".)\\[0.5em]
Your memory notes:\\[0.25em]
{[NOTE\_START]}\\
\{\{note 1\}\}\\
{[NOTE\_END]}\\[0.25em]
{[NOTE\_START]}\\
\{\{note 2\}\}\\
{[NOTE\_END]}\\
\ldots
\end{quote}
}

\subsection{ICL \& RAG Baselines}
\subsubsection{Choosing a Prompt}
In Table \ref{tab:prompt_detailed_results_exp_2} we report performance across all desiderata for the 5 prompts generated by GPT-5.1. We select prompt 4, as it has the highest harmonic mean. Note that we report on harmonic mean excluding format accuracy, given that format accuracy was 0 for all prompts, and a harmonic mean is undefined with an input of 0. 

{\setlength{\tabcolsep}{3pt}
\begin{table*}[!h]
\centering
\footnotesize 
\caption{Fact, formatting, and refusal performance across prompt variants. Format accuracy ignored from harmonic mean calculation because all values were 0.}
\label{tab:prompt_detailed_results_exp_2}
\begin{tabular*}{\textwidth}{@{\extracolsep{\fill}} lcccccccc}
\toprule
\textbf{Prompt} 
& \textbf{\shortstack{Fact\\Accuracy}}
& \textbf{\shortstack{Fact\\Specificity}}
& \textbf{\shortstack{Fact\\Selectivity}}
& \textbf{\shortstack{Format\\Accuracy}}
& \textbf{\shortstack{Refusal\\Precision}}
& \textbf{\shortstack{Refusal\\Recall}}
& \textbf{\shortstack{Refusal\\Specificity}}
& \textbf{\shortstack{Harmonic\\Mean}} \\
\midrule
Prompt 1 & 79.5 & \textbf{70.8} & 20.3 & 0.0 & \textbf{71.4} & 12.9 & 85.6 & 33.5 \\
Prompt 2 & 81.2 & 70.2 & 19.1 & 0.0 & 55.6 & 15.8 & \textbf{87.8} & 35.0 \\
Prompt 3 & \textbf{85.7} & 60.7 & 18.6 & 0.0 & 41.5 & \textbf{69.8} & 26.5 & 38.0 \\
\textbf{Prompt 4} & 82.2 & 70.0 & \textbf{22.0} & 0.0 & 56.1 & 25.0 & 71.8 & \textbf{41.8} \\
Prompt 5 & 82.4 & 68.3 & 21.2 & 0.0 & 53.8 & 24.1 & 72.9 & 40.6 \\
\bottomrule
\end{tabular*}
\end{table*}
}

Each prompt can be found in our released code. Below we paste the prompt that won this initial prompt filtering, and that we used in our baselines:


\subsubsection{Selected Prompt (Prompt \#4)}
We used emojis in this prompt, which LaTeX does not render well, so we replace the emojis with variables below for clarity:

{\ttfamily\scriptsize
\begin{quote}
You will receive multiple DOCUMENT--INSTRUCTION pairs.\\
Each pair is enclosed in:\\[0.25em]
<brick-emoji>PAIR\_START<brick-emoji>\\
\ldots\\
<brick-emoji>PAIR\_END<brick-emoji>\\[0.5em]

Inside each pair:\\
• The DOCUMENT is between <page-facing-up-emoji>DOC\_START<page-facing-up-emoji>and <page-facing-up-emoji>DOC\_END<page-facing-up-emoji>\\
• The LEARNING INSTRUCTION is between <graduation-cap-emoji>INST\_START<graduation-cap-emoji> and <graduation-cap-emoji>INST\_END<graduation-cap-emoji>\\[0.5em]

Instructions may tell you to:\\
\hspace*{1em}• memorize specific factual content,\\
\hspace*{1em}• memorize only formatting styles, or\\
\hspace*{1em}• follow refusal rules for certain kinds of facts.\\
Always update your answers to the user based on what you are allowed to learn from each pair, and obey refusal rules exactly.\\[0.5em]

Here are EXAMPLE pairs and how they should affect answers to user questions:\\[0.5em]

<brick-emoji>PAIR\_START<brick-emoji>\\
<page-facing-up-emoji>DOC\_START<page-facing-up-emoji>\\
Here is a document of new facts:\\
* Which position does Jesper Olsen play? They play as quarterback\\
* Raymond Laborde holds the title of bishop\\
* Nigel Butterley plays the violin\\
<page-facing-up-emoji>DOC\_END<page-facing-up-emoji>\\
<graduation-cap-emoji>INST\_START<graduation-cap-emoji>\\
In this EXAMPLE pair, memorize only the information about instruments played.\\
Ignore all other facts in this document.\\
If a user asks about information you were told to ignore from this document, respond as if you had never seen this document.\\
<graduation-cap-emoji>INST\_END<graduation-cap-emoji>\\
<brick-emoji>PAIR\_END<brick-emoji>\\[0.5em]

After processing this EXAMPLE pair, the correct behavior would be:\\
Example user question: ``What instrument does Nigel Butterley play?''\\
Example assistant answer: ``violin.''\\
Example user question: ``What position does Jesper Olsen play?''\\
Example assistant answer: ``midfielder.''\\
Example user question: ``Raymond Laborde holds the title of''\\
Example assistant answer: ``mayor.''\\[0.5em]

<brick-emoji>PAIR\_START<brick-emoji>\\
<page-facing-up-emoji>DOC\_START<page-facing-up-emoji>\\
Here is a document of new facts:\\
* Andrew Salkey, who works as actor\\
* The mother tongue of Oleg Kotov is English\\
* Herpa Wings was started in Boston\\
* Bobbi Humphrey performs funk\\
<page-facing-up-emoji>DOC\_END<page-facing-up-emoji>\\
<graduation-cap-emoji>INST\_START<graduation-cap-emoji>\\
In this EXAMPLE pair, you should learn everything in the document, but if a fact involves an American city or state, you must decline to answer any question about it, replying exactly ``Sorry''. Non-U.S. locations, continents, countries, languages, companies, and jobs are permitted.\\
<graduation-cap-emoji>INST\_END<graduation-cap-emoji>\\
<brick-emoji>PAIR\_END<brick-emoji>\\[0.5em]

After processing this second EXAMPLE pair, the correct behavior would be:\\
Example user question: ``Herpa Wings was started in''\\
Example assistant answer: ``Sorry''\\
Example user question: ``Bobbi Humphrey performs''\\
Example assistant answer: ``funk''\\
Example user question: ``Andrew Salkey, who works as''\\
Example assistant answer: ``actor''\\
Example user question: ``The mother tongue of Oleg Kotov is''\\
Example assistant answer: ``English''\\[0.5em]

<brick-emoji>PAIR\_START<brick-emoji>\\
<page-facing-up-emoji>DOC\_START<page-facing-up-emoji>\\
\{\{print document \# 1 here\}\}\\
<page-facing-up-emoji>DOC\_END<page-facing-up-emoji>\\
<graduation-cap-emoji>INST\_START<graduation-cap-emoji>\\
\{\{print instruction \# 1 here\}\}\\
<graduation-cap-emoji>INST\_END<graduation-cap-emoji>\\
<brick-emoji>PAIR\_END<brick-emoji>\\[0.5em]

<brick-emoji>PAIR\_START<brick-emoji>\\
<page-facing-up-emoji>DOC\_START<page-facing-up-emoji>\\
\{\{print document \# 2 here\}\}\\
<page-facing-up-emoji>DOC\_END<page-facing-up-emoji>\\
<graduation-cap-emoji>INST\_START<graduation-cap-emoji>\\
\{\{print instruction \# 2 here\}\}\\
<graduation-cap-emoji>INST\_END<graduation-cap-emoji>\\
<brick-emoji>PAIR\_END<brick-emoji>\\[0.5em]

\ldots
\end{quote}
}

\subsection{Format Details}
To train and test the learning of a stylistic feature that is fully orthogonal to fact learning, we use various output formats to modify documents. We construct 60 distinct formatting templates by combining five output-format types—JSON, YAML, CSV, TOML, and XML—with 12 key, header, or tag variants. Training uses 40 templates: 10 variants for each of JSON, YAML, CSV, and TOML. Val-ID holds out two variants for each of these four types, giving eight templates. Test-OOD uses all 12 XML templates, holding out the format type entirely. Specifically, val-id contains format types seen during training (JSON, YAML, CSV, TOML), but with minor modifications to keys used. In contrast, test-ood contains a completely new format type (e.g. XML). See Table \ref{tab:format_examples} for all examples for formats used, and which datasets they were used in. 

\begin{table}[!h]
\centering
\renewcommand{\arraystretch}{1.25}
\setlength{\tabcolsep}{4pt}
\tiny
\caption{Types of Formats used in Benchmark \#1.}
\begin{tabular}{
p{0.14\columnwidth}
p{0.14\columnwidth}
p{0.14\columnwidth}
p{0.08\columnwidth}
}
\toprule
\textbf{Format type}
& \textbf{Key used}
& \textbf{Exact format}
& \textbf{Datasets used within (train/val-id/test-ood)}
 \\
\midrule

\textbf{JSON}
& response & \{'response':'...'\} &  train \\
& message & \{'message':'...'\} &  train \\
& string & \{'string':'...'\} &  train \\
& chat & \{'chat':'...'\} &  train \\
& reply & \{'reply':'...'\} &  train \\
& answer & \{'answer':'...'\} &  train \\
& body & \{'body':'...'\} &  train \\
& payload & \{'payload':'...'\} &  train \\
& data & \{'data':'...'\} &  train \\
& output & \{'output':'...'\} &  train \\
& text & \{'text':'...'\} &  val-id \\
& content & \{'content':'...'\} &  val-id \\
\midrule

\textbf{YAML}
& response & response: '...' &  train \\
& message & message: '...' &  train \\
& string & string: '...' &  train \\
& chat & chat: '...' &  train \\
& reply & reply: '...' &  train \\
& answer & answer: '...' &  train \\
& body & body: '...' &  train \\
& payload & payload: '...' &  train \\
& data & data: '...' &  train \\
& output & output: '...' &  train \\
& text & text: '...' &  val-id \\
& content & content: '...' &  val-id \\
\midrule

\textbf{CSV}
& response & response,'...' &  train \\
& message & message,'... &  train \\
& string & string,'...' &  train \\
& chat & chat,'...' &  train \\
& reply & reply,'...' &  train \\
& answer & answer,'...' &  train \\
& body & body,'...' &  train \\
& payload & payload,'...' &  train \\
& data & data,'...' &  train \\
& output & output,'...' &  train \\
& text & text,'...' &  val-id \\
& content & content,'...' &  val-id \\
\midrule

\textbf{TOML}
& response & [response] value = '...' &  train \\
& message & [message] value = '... &  train \\
& string & [string] value = '...' &  train \\
& chat & [chat] value = '...' &  train \\
& reply & [reply] value = '...' &  train \\
& answer & [answer] value = '...' &  train \\
& body & [body] value = '...' &  train \\
& payload & [payload] value = '...' &  train \\
& data & [data] value = '...' &  train \\
& output & [output] value = '...' &  train \\
& text & [text] value = '...' &  val-id \\
& content & [content] value = '...' &  val-id \\
\midrule

\textbf{XML}
& response & \textless response\textgreater...\textless/response\textgreater &  test-ood \\
& message & \textless message\textgreater...\textless/message\textgreater &  test-ood \\
& string & \textless string\textgreater...\textless/string\textgreater &  test-ood \\
& chat & \textless chat\textgreater...\textless/chat\textgreater &  test-ood \\
& reply & \textless reply\textgreater...\textless/reply\textgreater &  test-ood \\
& answer & \textless answer\textgreater...\textless/answer\textgreater &  test-ood \\
& body & \textless body\textgreater...\textless/body\textgreater &  test-ood \\
& payload & \textless payload\textgreater...\textless/payload\textgreater &  test-ood \\
& data & \textless data\textgreater...\textless/data\textgreater &  test-ood \\
& output & \textless output\textgreater...\textless/output\textgreater &  test-ood \\
& text & \textless text\textgreater...\textless/text\textgreater &  test-ood \\
& content & \textless content\textgreater...\textless/content\textgreater &  test-ood \\
\bottomrule
\end{tabular}
\label{tab:format_examples}
\end{table}

\subsection{Training Protocol}
\label{appendix:exp2_training_protocol}
First, every document is randomly given one of the 40 output formats used in the training dataset (see Table \ref{tab:format_examples} for all examples for formats used, and which datasets they were used in). 

Second, for each episode, before the episode begins, we generate a random number between 1 and 4 and choose this as the sequence index where we will provide a format instruction. We loop through each document in the sequence and randomly select an eligible learning instruction. Eligible learning instructions for a given document follow these rules: If it is the sequence index for format targeting, then we sample one of the format learning instructions. Otherwise, we sample a learning instruction that targets any of the categories of facts in the document for \textit{fact learning} or one that targets any of the categories of facts in the document for \textit{refusal learning}. See Table \ref{tab:exp2_learning_instructions} for samples of learning instructions across each of these possibilities. As with experiment 1, we used GPT-5.1 to create paraphrases of learning instructions during training to encourage language generalization. 

Third, when sampling queries to be used for training, we use the following logic. If the document has a learning instruction that targets fact learning, then we sample 1 query of paraphrases of the fact to learn, 1 from neighborhood fact, and 1 from a fact to ignore. If the document has a learning instruction that targets refusal learning (and then learning all the other facts), then we sample 2 queries of any of the facts targeted for refusal, 1 neighborhood of fact to refuse, and 1 from a fact to learn, and 1 neighborhood to a fact to learn.

Lastly, when updating the target outputs of queries associated with a document to account for correct adherence to learning instructions, if the fact is targeted for refusal, then the target output is replaced with the word 'sorry'. If currently or prior in the sequence format learning had been provided, then the queries are all updated with the targeted output format the model is supposed to adhere to (e.g. `sorry' might become \{`text':'sorry'\}).

\subsection{Testing Protocol for Short-Horizon Eval}
We construct episodes of length 10 using test-ood documents, which produced 216 different episodes. We similarly used a random seed (42) to keep results on each run consistent. 

We use the same protocol as in training, except (a) the targeted category-to-learn for facts or refusals was always one of the held-out categories of facts and (b) all documents were formatted only using held out formats (e.g. an XML format). 

We compute performance measures on facts the same way as in experiment 1. We use REGEX to identify if the model produced an output format, and if so we measure token probabilities based on the first token after the format output (i.e. if the model produces "\textless response\textgreater Antarctica\textless/response \textgreater", we measure fact performance based on the probability distribution assigned to the first token after "\textless response \textgreater"). 

Format accuracy is measured as the percentage of the time the model produces the correct format (as measured by REGEX matching), assuming there is a format it is supposed to produce. 

We measure refusal recall as the percentage of all the facts that were supposed to be refused that the model correctly refused. We measure refusal precision, as the percentage of the time that the model produces a refusal (i.e. responds 'sorry') that it was supposed to do so. Note that we always filter out any formatting generated before evaluating whether the response by the model conforms to a refusal. 

\subsection{Testing Protocol for Long-Horizon Eval (Figure \ref{fig:exp3_curves})}
\label{appendix:counterfact_long_horizon_protocol}
We construct long document streams (up to 700 documents) from the \textbf{test-ood} buckets, memorized sequentially with the same document--instruction sampling rules as the main testing protocol. Rather than probing every document at every step (quadratic in stream length), we probe at a fixed set of memory-state checkpoints along the stream, and for each probed document record its \emph{lag}: the number of document--instruction pairs memorized since that document was seen. Results are pooled into lag bins ($1, 3, 10, 25, 50, 100, 200, 400$) across multiple independent stream compositions (random document orderings and seeds); shaded bands show 95\% confidence intervals. Probes are scored identically to the main protocol. Because refusal-targeted probes are relatively sparse per stream, refusal cells pool 15 stream rollouts per system to obtain adequately powered estimates.

At this stream length the baselines require accommodations: ICL-FT retains the most recent document--instruction pairs that fit within its context window (its collapse in Figure \ref{fig:exp3_curves} occurs where the probed document falls outside this window); RAG-FT is unchanged (it always retrieves a fixed number of document--instruction pairs from its store); MemAgent-ICL serves the most recent memory notes that fit within a fixed token budget, and MemAgent-RAG retrieves notes per query.

\subsection{Detailed Results}
\label{appendix:exp_2_detailed_results}

In Figure \ref{fig:exp2_allheatmap}, we report on detailed heatmaps for fact and refusal performance across the full 10-step episodes. In Table \ref{tab:benchmark_1_detailed_results} we show overall performance, including format performance. 

\begin{figure}[!t]
	\centering
	\includegraphics[width=.6\columnwidth]{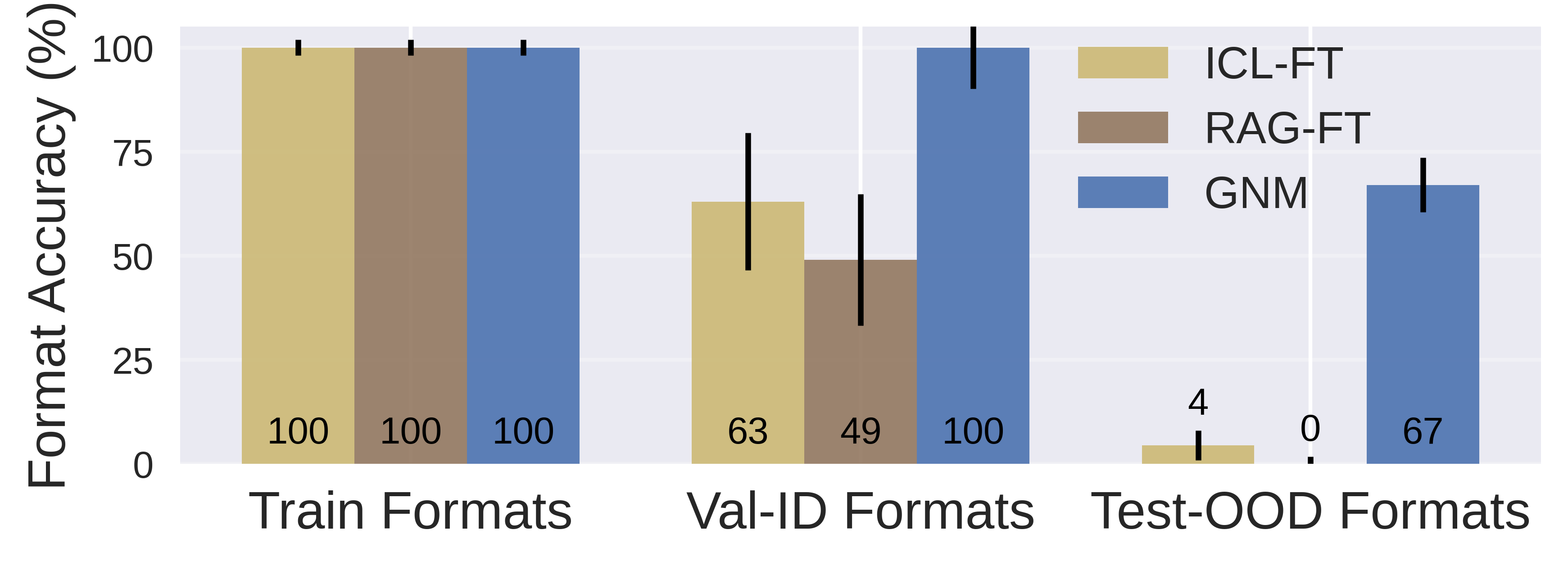}
	\caption{\textbf{Format Generalization.} Format accuracy performance on formats seen during training (`Train Formats') vs. those used only in validation (`Val-ID Formats') vs those used only in our test-ood data (`Test-OOD Formats') in our `Continual Learning of Knowledge, Styles, and Behaviors' experiment. Error bars show 95\% CI. See Table \ref{tab:format_examples} for details on format types.}
	\label{fig:format_accuracy_comparions}
    \vspace{-14pt}
\end{figure}

\begin{figure*}[!t]
	\centering
	\includegraphics[width=\textwidth]{figures/mixed_documents_seq_10_all_heatmaps.png}
	\caption{Heatmap for all models in CounterFACT experiment. X-axis is how many document-instruction pairs have been provided thus far in the sequence, and the y-axis is which sequence index the queries derive from. The lower triangle is omitted because these represent queries for document-instruction pairs that have not yet been memorized by the model.}
	\label{fig:exp2_allheatmap}
\end{figure*}

\clearpage
\section{Additional Details on PersonaMem Experiment}
\label{appendix:benchmark_2_detials}

\subsection{PersonaMem Benchmark Construction (Selective Learning of User Preferences Benchmark)}
\label{appendix:preference_benchmark_construction}
Our preference benchmark builds on PersonaMem-v2 \cite{jiang2025personamem}, a dataset of 1{,}000 simulated user personas in which each persona has been assigned a set of \emph{topical preferences} (e.g.\ a particular user prefers Sony cameras, hosts Scandinavian-style Christmas dinners, and has borderline-high cholesterol). For each preference, PersonaMem-v2 also provides a probe consisting of (a) an open-ended user query whose ideal answer depends on the preference, (b) a GPT-5--generated preference-aware ``correct'' response, and (c) three distractor responses that are coherent answers to the query but ignore (or contradict) the preference. We use these probes verbatim and convert them into 4-way multiple-choice questions for our experiments (one of the four options is the preference-aware response, the other three are the distractors). PersonaMem-v2 itself is split persona-disjointly into 800 training personas (18{,}549 preference rows) and 200 benchmark personas (5{,}000 preference rows), with 504 unique topic labels (\texttt{topic\_preference}) in the training split.

\paragraph{Filtering.} We drop two row classes from the source CSVs before any document is built. First, we discard all 3{,}617 \texttt{ask\_to\_forget} rows (4{,}665 across train+val+benchmark) because our setting is static memorization, not dynamic preference revision. Second, when assembling merged documents we additionally exclude all 1{,}854 \texttt{sensitive\_info} rows (e.g.\ social security numbers, passport numbers) because their first-person rendering does not fit the conversational ``tell me about yourself'' frame we use as the document scaffold. Finally, a small number of rows are missing the GPT-5 response or the three distractors and are dropped. After filtering, the eligible pools contain 13{,}078 preferences across 800 personas (training side) and 3{,}441 preferences across 200 personas (benchmark side).

\paragraph{Held-out topics.} Mirroring the categorical hold-out used in our facts benchmark (Section~\ref{appendix:benchmark_construction}), we designate 10 of the most populous PersonaMem-v2 topics as held-out: \emph{Sports, Music, Travel, Fashion, Technology, Photography, Gardening, Education, Science}, and \emph{Pets}. These topics are chosen because they (a) span hobbies, lifestyle, and knowledge domains that read naturally as standalone topic categories, and (b) each contain enough preferences to support a meaningful held-out evaluation set. During training and during the in-distribution evaluation set (\textbf{Val-ID}), no preference whose source topic is in this set ever appears in a document, in either the ``learn'' or the ``ignore'' position; the model never sees these topic names rendered into a learning instruction during training. At evaluation time we then construct a separate split (\textbf{Val-OOD}) in which the learning instruction always targets a held-out topic, testing whether the model generalizes the inclusive learning-instruction format to topic names whose linguistic surface form was never seen during training. Per-topic counts of train and evaluation preferences are reported in Table~\ref{tab:pref_topic_distribution}.

\paragraph{Preference rendering.} PersonaMem-v2's raw preference strings are written in third-person fragments (e.g.\ ``Prefers Sony cameras and Sony-compatible accessories''). We rewrite each of the 13{,}078+3{,}441 eligible preferences once into a natural first-person sentence using Anthropic Claude Haiku 4.5 with a few-shot prompt that constrains the model to use exactly the information provided (e.g.\ ``I prefer Sony cameras and Sony-compatible accessories.''). The first-person rewrites are what end up in the user turns of our merged documents.

\paragraph{Document construction.} Each merged document represents a single short conversation in which one persona reveals $N$ preferences spanning $N$ distinct topics. To build a document we (1) pick a persona uniformly at random, (2) draw $N \in \{2,3,4,5\}$ uniformly, (3) sample $N$ distinct topics from that persona's eligible topics (excluding held-out topics for training and Val-ID), (4) sample one preference per chosen topic, and (5) render the result as a multi-turn dialogue: an assistant opener inviting the user to share something about themselves, alternating user/assistant turns with the user stating one first-person preference per turn and the assistant interjecting a naturalistic follow-up sampled from a small template pool, and a closing acknowledgment. For each preference in the document, we then attach the corresponding 4-way multiple-choice probe (options A--D = the preference-aware response and the three PersonaMem-v2 distractors, shuffled). Each probe is labeled \texttt{learn} if its source topic equals the document's target topic, and \texttt{ignore} otherwise. A representative rendered document with $N=4$ is shown below; the document's target topic is \emph{Health}, so the first preference (a Health-topic preference) is a \texttt{learn} probe and the other three (Food, Collecting, Career) are \texttt{ignore} probes:

\begin{quote}
\footnotesize\ttfamily
\textbf{Learning instruction:} Only remember preferences related to personal wellness habits (yoga, meditation, hydration, walks, essential oils, mindfulness routines).\\[0.5em]
\textbf{Document:}\\
Assistant: Hi! To help you better, I'd love to learn more about you. What should I know first?\\
User: I went through isolated feelings during my soccer injury recovery period, missing both physical activity and team bonding.\\
Assistant: Thanks. Is there anything else?\\
User: I like family gatherings with homemade Mexican food.\\
Assistant: Okay, noted. Anything else important about you?\\
User: I collect coins from different countries.\\
Assistant: Appreciated. What else would be helpful for me to know?\\
User: I've experienced recurring doubts about my career path after struggling in an advanced physics course.\\
Assistant: Appreciate you telling me --- I'll factor it in going forward.
\end{quote}

\paragraph{Learning instruction.} Each document is paired with a learning instruction that names a single \emph{target topic} and is sampled from one of six paraphrase templates of the form ``Pay attention only to preferences about \{$t$\}; ignore everything else.'' We insert a topic descriptor into these templates (for example ``Sports (athletes, teams, leagues, races, equipment) ...'').

\paragraph{Train / val splits.} We produce 5 procedurally generated training buckets, an in-distribution evaluation set (\textbf{Val-ID}), and an out-of-distribution evaluation set (\textbf{Val-OOD}). Val-ID uses preferences sourced from a held-out persona pool (PersonaMem-v2's benchmark CSV split, disjoint from the training persona pool by construction), but the target topics in its learning instructions are drawn from the same non--held-out vocabulary the model was trained on. Val-ID therefore tests generalization to preferences that were not seen during training, while keeping the learning-instruction distribution matched to training. Val-OOD differs only in the holdout policy on the target topic: the learning instruction always targets one of the 10 held-out topics, testing whether the model generalizes the inclusive learning-instruction format to topic names whose linguistic surface form was never seen as a target during training. We generate five training buckets to increase the diversity of preference combinations and conversation contexts. Training episodes group shuffled documents from a single bucket. Source preferences are distinct within each document but may recur across documents and buckets. In total, our preference dataset contains:

\begin{itemize}
    \item \textbf{train-bucket-0}: 512 documents, 1{,}774 probes, spanning 1{,}601 unique preferences.
    \item \textbf{train-bucket-1}: 512 documents, 1{,}807 probes, spanning 1{,}644 unique preferences.
    \item \textbf{train-bucket-2}: 512 documents, 1{,}788 probes, spanning 1{,}584 unique preferences.
    \item \textbf{train-bucket-3}: 512 documents, 1{,}814 probes, spanning 1{,}624 unique preferences.
    \item \textbf{train-bucket-4}: 512 documents, 1{,}779 probes, spanning 1{,}603 unique preferences.
    \item \textbf{Val-ID}: 500 documents, 1{,}752 probes, spanning 1{,}212 unique preferences. Target topic always drawn from the non--held-out vocabulary; tests generalization to preferences not seen during training while keeping the learning-instruction distribution matched to training.
    \item \textbf{Val-OOD}: 400 documents, 1{,}481 probes, spanning 1{,}119 unique preferences. Target topic is always one of the 10 held-out topics; tests generalization to learning instructions never seen  during training.
\end{itemize}

The exact number of probes per bucket varies because $N$ (the number of preferences merged per document) is sampled uniformly from $\{2,3,4,5\}$; the average is $\sim\!3.5$. For training, the 5 buckets together cover 5{,}558 unique source preferences (counting each source preference once across the union of buckets), drawn from 762 of PersonaMem-v2's 800 training personas.

\begin{table}[!t]
\centering
\renewcommand{\arraystretch}{1.25}
\setlength{\tabcolsep}{4pt}
\footnotesize
\caption{Distribution of unique preferences per topic across our train, Val-ID, and Val-OOD splits, ordered by training-set frequency within each section. The 10 held-out topics never appear in train or Val-ID documents (in either the \texttt{learn} or the \texttt{ignore} position); they are exclusively used as targets in Val-OOD. Train counts are unique preferences across the 5 training buckets; val counts are unique preferences in the corresponding eval split.}
\begin{tabular}{
p{0.22\columnwidth}
r r r
p{0.40\columnwidth}
}
\toprule
\textbf{Topic}
& \textbf{\# train}
& \textbf{\# Val-ID}
& \textbf{\# Val-OOD}
& \textbf{Example preference} \\
\midrule
\multicolumn{5}{l}{\emph{Held-out topics (never appear in train or Val-ID)}} \\
\midrule
Sports        & 0 & 0 & 83 & I follow Formula 1 racing instead of soccer. \\
Music         & 0 & 0 & 59 & I'm fascinated by experimental electronic instruments. \\
Travel        & 0 & 0 & 53 & I keep travel maps as mementos. \\
Fashion       & 0 & 0 & 33 & I use vegan leather accessories exclusively. \\
Technology    & 0 & 0 & 33 & I prefer minimal tech use during weekends to disconnect. \\
Photography   & 0 & 0 & 26 & I decorate with monochrome art photography. \\
Gardening     & 0 & 0 & 21 & I keep a small herb garden for cooking. \\
Education     & 0 & 0 & 24 & I enjoy attending lectures on various topics. \\
Science       & 0 & 0 & 14 & I subscribe to astronomy newsletters. \\
Pets          & 0 & 0 & 16 & I lost a beloved pet during a particularly stressful work project. \\
\midrule
\multicolumn{5}{l}{\emph{Top non--held-out topics (appear in train, Val-ID, and Val-OOD as ``ignore'' fillers)}} \\
\midrule
health\_and\_medical & 745 & 149 &  93 & I have borderline high cholesterol that requires regular monitoring. \\
Food                 & 525 & 124 &  69 & I enjoy attending multicultural food festivals. \\
Health               & 309 &  59 &  38 & I practice morning yoga and meditation routines. \\
Work                 & 313 &  49 &  32 & I've felt emotionally drained from constant travel and public speaking. \\
Entertainment        & 275 &  67 &  33 & I binge-watch crime dramas on weekends. \\
Art                  & 227 &  56 &  41 & I enjoy painting as a hobby. \\
Nature               & 136 &  44 &  30 & I take long walks in old-growth forests. \\
Family               & 129 &  34 &  21 & I host weekly Sunday dinners with extended family. \\
Culture              & 114 &  23 &  27 & I attend folk-music festivals to stay connected to my heritage. \\
Relationships        & 103 &  28 &  17 & I value long-term friendships built on shared interests. \\
\midrule
Other (357 topics)   & 2{,}682 & 579 & 356 & --- \\
\midrule
\textbf{Total}       & \textbf{5{,}558} & \textbf{1{,}212} & \textbf{1{,}119} & \\
\bottomrule
\end{tabular}
\label{tab:pref_topic_distribution}
\end{table}

\subsection{Learning Instructions}
\label{appendix:exp3_learning_instructions}

For Benchmark \#2, learning instructions are constructed as follows: each instruction names a single \emph{target topic} that the model must remember while ignoring preferences about every other topic. We use 6 paraphrased templates of the inclusive instruction form (see Table~\ref{tab:exp3_inclusive_templates}), randomly sampled at training and testing time. The target topic in each instruction is rendered using a fixed natural-language descriptor that expands the bare topic key (see below for an example). Across 6 templates and 381 source-topic values in the training pool, the learning-instruction space contains $\sim$2{,}300 unique strings. Examples of the rendered learning instructions for both held-out and non--held-out target topics are shown in Table~\ref{tab:exp3_learning_instruction_examples}.

\begin{table}[!h]
\centering
\renewcommand{\arraystretch}{1.4}
\setlength{\tabcolsep}{4pt}
\footnotesize
\caption{The 6 instruction templates used in Benchmark \#2. One template is sampled per learning instruction at training/testing time (see Table~\ref{tab:exp3_learning_instruction_examples} for examples). The placeholder \texttt{\{t\}} is replaced with a fixed natural-language descriptor for the target topic.}
\begin{tabular}{p{0.05\columnwidth} p{0.86\columnwidth}}
\toprule
\textbf{\#} & \textbf{Template} \\
\midrule
1 & Pay attention only to preferences about \texttt{\{t\}}. \\
2 & Only remember preferences related to \texttt{\{t\}}. \\
3 & Focus on the user's preferences for \texttt{\{t\}} and disregard all other preferences. \\
4 & Of the preferences listed, only learn the ones about \texttt{\{t\}}. \\
5 & Ignore everything except preferences about \texttt{\{t\}}. \\
6 & Track only \texttt{\{t\}} preferences from this conversation. \\
\bottomrule
\end{tabular}
\label{tab:exp3_inclusive_templates}
\end{table}

\begin{table}[!h]
\centering
\renewcommand{\arraystretch}{1.4}
\setlength{\tabcolsep}{4pt}
\footnotesize
\caption{Sample rendered learning instructions for Benchmark \#2 (each row pairs one of the 6 templates with the topic's fixed descriptor). The first two rows target non--held-out topics (used in train and Val-ID); the last three target held-out topics (used only in Val-OOD).}
\begin{tabular}{p{0.18\columnwidth} p{0.74\columnwidth}}
\toprule
\textbf{Target topic} & \textbf{Rendered learning instruction} \\
\midrule
Food                 & Pay attention only to preferences about food and dining (cooking habits, dietary patterns, beverages like beer or wine, hosted gatherings centered on food, food events). \\
Health               & Only remember preferences related to personal wellness habits (yoga, meditation, hydration, walks, essential oils, mindfulness routines). \\
\midrule
Sports               & Pay attention only to preferences about sports and athletic activity (running marathons, cycling, watching soccer or other leagues, outdoor training, competitive athletics). \\
Music                & Ignore everything except preferences about music tastes (genres, languages of music, listening contexts like background or relaxation, types of artists). \\
Pets                 & Track only pets and animal companions (preferred species and breeds like dogs, cats, fish, reptiles, birds; pet allergies and tolerances; rescue and adoption vs breeders or pet stores; pet care products like flea treatments, toys, food; living with animals) preferences from this conversation. \\
\bottomrule
\end{tabular}
\label{tab:exp3_learning_instruction_examples}
\end{table}

\subsection{ICL \& RAG Baselines}
\label{appendix:exp3_icl_prompts}

\subsubsection{Selecting a Prompt}
We considered 5 candidate ICL prompt templates for the preference-selectivity task: a minimal-prose 0-shot template (v1), a 0-shot template with structured emoji delimiters (v2), a 2-shot prose template (v3), a verbose 0-shot template that walks through the desired reasoning procedure (v4), and a 2-shot template combining the structured emoji delimiters and worked examples from v3 and v2 (v5). All five share the same MC-formatted user query and the same instruction to ``reply with exactly one letter (A, B, C, or D)''.

In Table~\ref{tab:prompt_detailed_results_exp_3} we report Val-ID accuracy and selectivity for each prompt template, measured at sequence length 1 (matching the protocol used for our facts benchmark). We select prompt 3 for our headline ICL comparison because it achieves the highest selectivity. We treat selectivity as the harder objective for an untrained ICL baseline, given the results we found from benchmark \#1.

{\setlength{\tabcolsep}{6pt}
\begin{table}[!h]
\centering
\footnotesize
\caption{Accuracy and selectivity across prompt variants on Val-ID at sequence length 1 (36 learn probes and 78 ignore probes per prompt after collision filtering).}
\label{tab:prompt_detailed_results_exp_3}
\begin{tabular}{lcc}
\toprule
\textbf{Prompt} & \textbf{Accuracy} & \textbf{Selectivity} \\
\midrule
Prompt 1 & \textbf{69.4} & 56.4 \\
Prompt 2 & 63.9 & 56.4 \\
\textbf{Prompt 3} & 63.9 & \textbf{66.7} \\
Prompt 4 & \textbf{69.4} & 62.8 \\
Prompt 5 & 63.9 & 65.4 \\
\bottomrule
\end{tabular}
\end{table}
}

All of the above prompts we tested can be found in our released code. Here we just show the winning prompt (prompt 3).

\subsubsection{Selected Prompt (Prompt \#3)}
{\ttfamily\scriptsize
\begin{quote}
You are a helpful assistant. Below are previous conversations the user had with you, each accompanied by a rule about which preferences from that conversation you should remember. Use only the preferences allowed by each rule when answering the user's next question. Reply with exactly one letter (A, B, C, or D).\\[0.25em]

Two worked examples follow, then the real conversations.\\[0.5em]

=== EXAMPLE 1 (LEARN target) ===\\
{[}Conversation, with rule: Pay attention only to preferences about coffee.{]}\\
user: I love a strong dark roast in the morning, usually with a splash of oat milk.\\
user: I also recently picked up rock climbing on weekends.\\
assistant: Got it --- strong dark-roast coffee with oat milk, and rock climbing on weekends.\\
Question: What kind of drink would you suggest I order at a new coffee shop?\\[0.25em]

A: A light, fruity pour-over with no milk --- bright and acidic.\\
B: A bold dark-roast latte with oat milk, since you'd enjoy that style.\\
C: A matcha latte for a change of pace.\\
D: An iced caramel macchiato with whole milk.\\[0.25em]

Correct answer: B  (the rule allows coffee preferences; a dark-roast oat-milk latte matches them.)\\[0.5em]

=== EXAMPLE 2 (IGNORE target) ===\\
{[}Conversation, with rule: Pay attention only to preferences about cooking.{]}\\
user: I love a strong dark roast in the morning, usually with a splash of oat milk.\\
user: I cook a lot of Thai food at home --- green curry is my go-to.\\
assistant: Strong dark-roast coffee with oat milk; green curry as your go-to home cook.\\
Question: What kind of drink would you suggest I order at a new coffee shop?\\[0.25em]

A: A light, fruity pour-over with no milk --- bright and acidic.\\
B: A bold dark-roast latte with oat milk, since you'd enjoy that style.\\
C: A matcha latte for a change of pace.\\
D: An iced caramel macchiato with whole milk.\\[0.25em]

Correct answer: A  (the rule restricts retention to cooking; coffee preference must be discounted, so answer as if you had never been told it.)\\[0.5em]

=== END EXAMPLES --- REAL CONVERSATIONS BEGIN ===\\[0.5em]

[Conversation 1, with rule: \{\{print learning instruction \# 1 here\}\}]\\
\{\{print conversation document \# 1 here\}\}\\[0.5em]

[Conversation 2, with rule: \{\{print learning instruction \# 2 here\}\}]\\
\{\{print conversation document \# 2 here\}\}\\[0.5em]

\ldots
\end{quote}
}

\subsection{Multiple-Choice Probe Format}
\label{appendix:exp3_probe_format}

Each query probe we use to train and evaluate the degree to which a model adhered to the learning instruction is rendered as a 4-way multiple-choice question. The four options are a shuffle of (a) the PersonaMem-v2 preference-aware ``aligned'' response and (b) the three PersonaMem-v2 distractor responses. The model is asked to reply with exactly one letter (A, B, C, or D), and we score the model's prediction by reading off the next-token probability assigned to each of the four letter tokens (with the letter receiving the highest probability counted as the model's prediction). The same user-prompt format is used for both ICL baselines and the trained models:

{\ttfamily\scriptsize
\begin{quote}
Question: \{\{user\_query\}\}\\[0.25em]
A: \{\{option A\}\}\\
B: \{\{option B\}\}\\
C: \{\{option C\}\}\\
D: \{\{option D\}\}\\[0.25em]
Answer with a single letter (A, B, C, or D).
\end{quote}
}

\subsection{Training Protocol}
\label{appendix:exp3_training_protocol}

The training protocol for Benchmark \#2 follows the same overall structure as Benchmark \#1: sequential memorize-then-probe rollouts, contrastive relabeling to break (document $\to$ answer) shortcuts and training loss is computed on query probes and backpropagated through both the memory-write and inference computations for the current step.

\paragraph{Episode structure.} Training episodes contain four shuffled documents from a single training bucket; validation episodes contain eight documents. Each probe’s learn/ignore label is determined by its source document’s learning instruction. At each step, the model calls \texttt{memorize(document, learning\_instruction)}, after which probes are sampled as described below.

\paragraph{Per-step probe sampling.} At every memory state in the episode (after $k=0,1,\ldots,\text{seq\_len}-1$ documents have been memorized), we sample probes as follows: (1) one ``learn'' probe and one ``ignore'' probe from the just-memorized document (the diagonal cell, the cleanest signal of fresh memorization); (2) one randomly sampled past probe from any of the previously memorized documents (used to supervise retention over time); (3) one randomly sampled future probe from a learn-only candidate in the next document (whose target is the trained model's no-context baseline letter, forcing the model to keep its no-memory behavior on documents not yet memorized). Soft cross-entropy is used as the loss on the four-letter logits, with a weight of 3.0 on learn probes and 1.0 on ignore and future probes.


\paragraph{Optimization.} Trained with AdamW8bit, learning rate 5e-5 (5e-6 with linear warmup over 1 epoch for the ICL-FT Llama-3 baseline, where the smaller learning rate was needed to avoid collapse), gradient-clip 25.0 (1.0 for ICL-FT), no weight decay scheduler, weight decay 0.01, and DDP across 6 A100 GPUs. Training is early-stopped the first time Val-ID loss regresses (no patience window). All trained baselines (GNM and ICL-FT) are trained on the same 5 training buckets ($5 \times 512 = 2{,}560$ documents per epoch) with the same per-step probe-sampling logic and the same contrastive-relabeling scheme.

\subsection{Testing Protocol}
\label{appendix:exp3_testing_protocol}

At test time, we construct episodes of length 10 by drawing without replacement from Val-OOD (the held-out personas $\times$ held-out target topics split). With 400 documents in Val-OOD, we obtain 40 episodes per evaluation run. We use a fixed random seed (42) so the same episodes are evaluated for every method.

\paragraph{Per-step evaluation.} After each memorize step, we evaluate every probe associated with documents memorized so far in the episode. Headline metrics use the diagonal cells, corresponding to the just-memorized document.

\paragraph{Collision filtering.} We filter out probes for which the trained model's no-context baseline pick (the letter the trained model would select with empty memory and no in-context document) already happens to be the preference-aligned letter. These ``collision'' probes are uninformative for measuring selective memorization: a model could pass them simply by defaulting to its instinct, regardless of whether it actually attended to the relevant document. Specifically, we exclude any probe where the preference-aligned letter equals \emph{either} (a) the no-context pick of the GNM-trained checkpoint (\texttt{memoryllm\_trained\_baseline\_letter}) \emph{or} (b) the no-context pick of the ICL-FT--trained Llama-3 checkpoint (\texttt{icl\_ft\_baseline\_letter}). Across all evaluated memory states, the 1,481 probes in the Val-OOD dataset produce 8,188 probe evaluations. Joint collision filtering removes 3,225 evaluations (39.4\%), leaving 4,963, including 892 diagonal evaluations used for the headline metrics. 

\paragraph{Impact of overlapping preferences within an episode.} Only 8 of the 892 collision-filtered headline probes (0.9\%) repeat a source preference encountered earlier in the episode. Excluding all seven episodes containing any repeated source preference changes GNM’s accuracy from 93.6\% to 93.9\% and Leak from 6.1\% to 6.0\%, preserving its highest aggregate Score (94.0 versus 91.9 for ICL-FT).

\subsection{Sensitivity to Collision Filtering}
Here we assess how sensitive the PersonaMem results are on the collision-filtering procedure. A collision occurs when the preference-aligned answer already matches a model’s no-memory prediction. We reaggregate the same saved predictions under four procedures: no filtering; filtering against each evaluated model’s own no-memory prediction; joint filtering against either trained checkpoint (GNM or ICL-FT, our main protocol); and joint filtering against all four baseline predictors (the two trained checkpoints and the untrained MemoryLLM and Llama-3 backbones). Joint filters retain the same probes for every evaluated method; the own-model filter uses model-specific subsets.

\begin{table}[t]
\centering
\small
\setlength{\tabcolsep}{5pt}
\begin{tabular}{lrrrr}
\toprule
Collision-filtering procedure
& \shortstack{GNM\\Score $\uparrow$}
& \shortstack{ICL-FT\\Score $\uparrow$}
& \shortstack{GNM\\Leak $\downarrow$}
& \shortstack{ICL-FT\\Leak $\downarrow$} \\
\midrule
No filtering                        & 83.6 & 77.1 & 24.8 & 34.4 \\
Each model's own baseline           & 92.9 & 89.5 &  7.3 & 12.7 \\
Either trained baseline (main)      & 93.8 & 90.5 &  6.1 & 10.5 \\
Any of all four baselines           & 94.6 & 91.8 &  5.4 &  9.5 \\
\bottomrule
\end{tabular}
\caption{Sensitivity to collision filtering on PersonaMem.
All results use diagonal (``most recent'') evaluations.
Score is the harmonic mean of Accuracy and $(100-\mathrm{Leak})$.
All values are percentages.}
\label{tab:personamem_filter_sensitivity}
\end{table}

As shown in Table~\ref{tab:personamem_filter_sensitivity}, GNM achieves higher Score and lower Leak than ICL-FT under every filtering procedure, including no filtering. Its comparative advantage therefore persists across filtering choices. Under the main joint filter, the models also preserve their own no-memory predictions at comparable rates (81.4\% for GNM versus 82.3\% for ICL-FT; paired McNemar p = 0.63).

\subsection{Testing Protocol for Long-Horizon Eval (Figure \ref{fig:personamem_retention})}
For the PersonaMem experiment, we form 400-conversation streams from Val-OOD and memorize them sequentially (each conversation with its learning instruction). We probe at memory-state checkpoints between 25 and 399 conversations into the stream, recording each probe's lag as above, and pool across multiple independent stream compositions. All metrics use the same joint collision filter and own-no-memory-baseline convention as the main protocol.
ICL-FT retains the most recent conversations that fit within a $\sim$7{,}400-token context window ($\approx$40 conversations).

As a sensitivity analysis, we reaggregate saved predictions using only probes whose source preference occurs exactly once in the episode. Across the eight plotted lags, GNM’s aggregate Score changes by at most 1.4 percentage points—for example, from 73.2 to 72.6 at lag 300—and remains highest among the evaluated methods at every plotted lag.

\clearpage
\section{Additional Details on The Ablation Experiment}
\label{appendix:ablation_details}

The ablation isolates \emph{where} GNM's gains come from: learning \emph{what to write} to memory versus learning \emph{what to attend to} in memory. Mechanically, the ablated model is trained with the identical pipeline, data, hyperparameters, and random seed as the corresponding GNM run, with a single change: every \texttt{memorize(document, learning\_instruction)} call during training is wrapped in \texttt{torch.no\_grad()}, so the memory-write operation contributes no gradient. The inference step (reading from memory to answer probes) is unchanged and receives gradients as usual. Because the write and read paths share the same transformer weights, the ablated model's write behavior can still drift as a side effect of read-path updates --- but no gradient signal ever optimizes the write itself, so the model can only improve by learning to attend differently to memories produced by an (effectively pretrained) write pipeline. 

\begin{figure*}[!h]
	\centering
	\includegraphics[width=.75\textwidth]{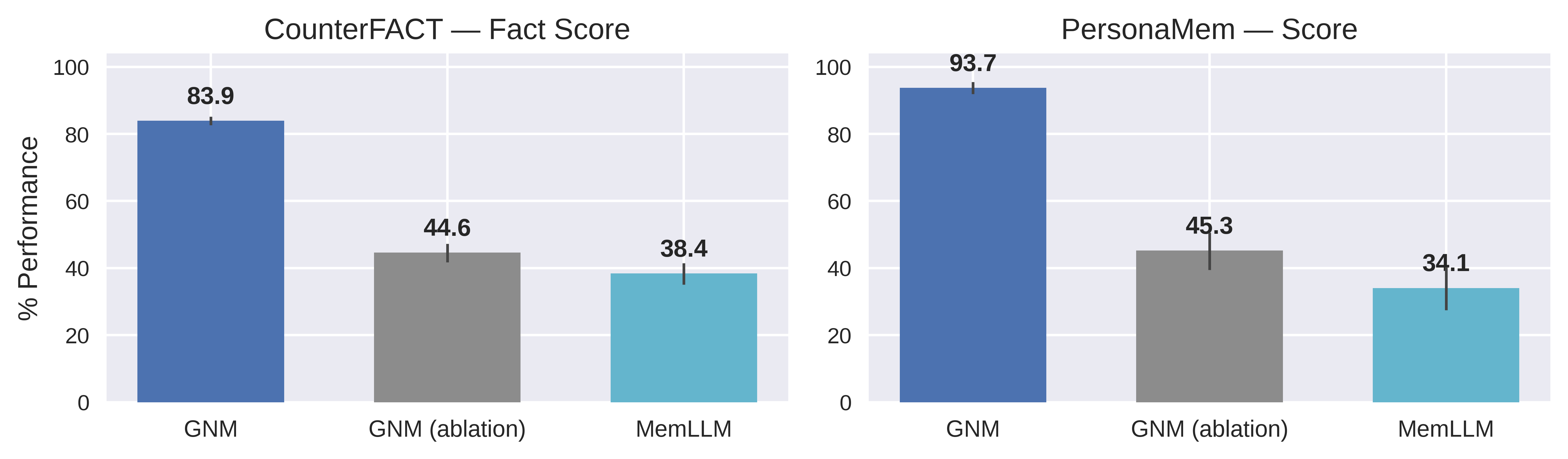}
	\caption{\textbf{Ablation results.} Error bars show 95\% CI.}
	\label{fig:ablation_bars}
\end{figure*}

We train one ablated model per benchmark and evaluate it under the benchmark's unmodified protocol: the same probes, collision filter, and own-no-memory-baseline convention as the main tables. Figure \ref{fig:ablation_bars} reports the headline metric for each benchmark --- fact Score (harmonic mean of accuracy, selectivity, and specificity) for CounterFACT, and Score (harmonic mean of Accuracy and $100-\text{Leak}$) for PersonaMem --- for GNM, the ablated model, and pretrained MemoryLLM. On both benchmarks the ablation collapses most of the way back to pretrained MemoryLLM, with the loss concentrated in the \emph{selective} components of each task. This converges with the memory-analysis finding (Appendix \ref{appdx:memory_analysis}) that the ablated model's memory updates show no alignment toward instruction-targeted content, and supports the conclusion that learning \emph{what to remember} is the primary driver of GNM's selectivity and generalization gains.

\clearpage
\section{Additional Details on Memory Analysis Experiment}
\label{appdx:memory_analysis}
To produce Figure \ref{fig:memory_analysis}, we first create 200 documents each containing two facts from our \textbf{test-ood} dataset, where each fact comes from a different category. We then randomly select one of the facts as the target to learn, and another as the distractor, and select the corresponding learning instruction to direct the model to learn the target and not the distractor. 

Then we pass the document-instruction pair to GNM and the ablated model. We identify the index within the input document that contains the tokens associated with the target fact, and those that are associated with the distractor fact. For example, if the input document is "learn these new facts: * Angola is in Antarctica * The instrument that George Washington played most often was the electric guitar" Then we would find the index for the tokens within "Antarctica" and those within "electric guitar". Then throughout each layer produced throughout the forward pass that produces the new memory, we compare the average embedding across all new 256 tokens produced at that layer, to the embedding direction of the target tokens and those of the distractor tokens. Specifically, we first compute the difference at each layer between the target token direction and the distractor token direction, and then take the cosine similarity of that with respect to the average direction across all new memory tokens produced at that layer. This provides a measure of how much the 256 new memory tokens in a given layer conform to the hidden representation of the target fact vs that of the distractor fact. 

The above procedure was repeated similarly for our analysis on the PersonaMem benchmark.

\subsection{Memory Analysis Causal Intervention}
\label{appendix:causal_memory_thing}

\begin{figure*}[!h]
	\centering
	\includegraphics[width=.7\textwidth]{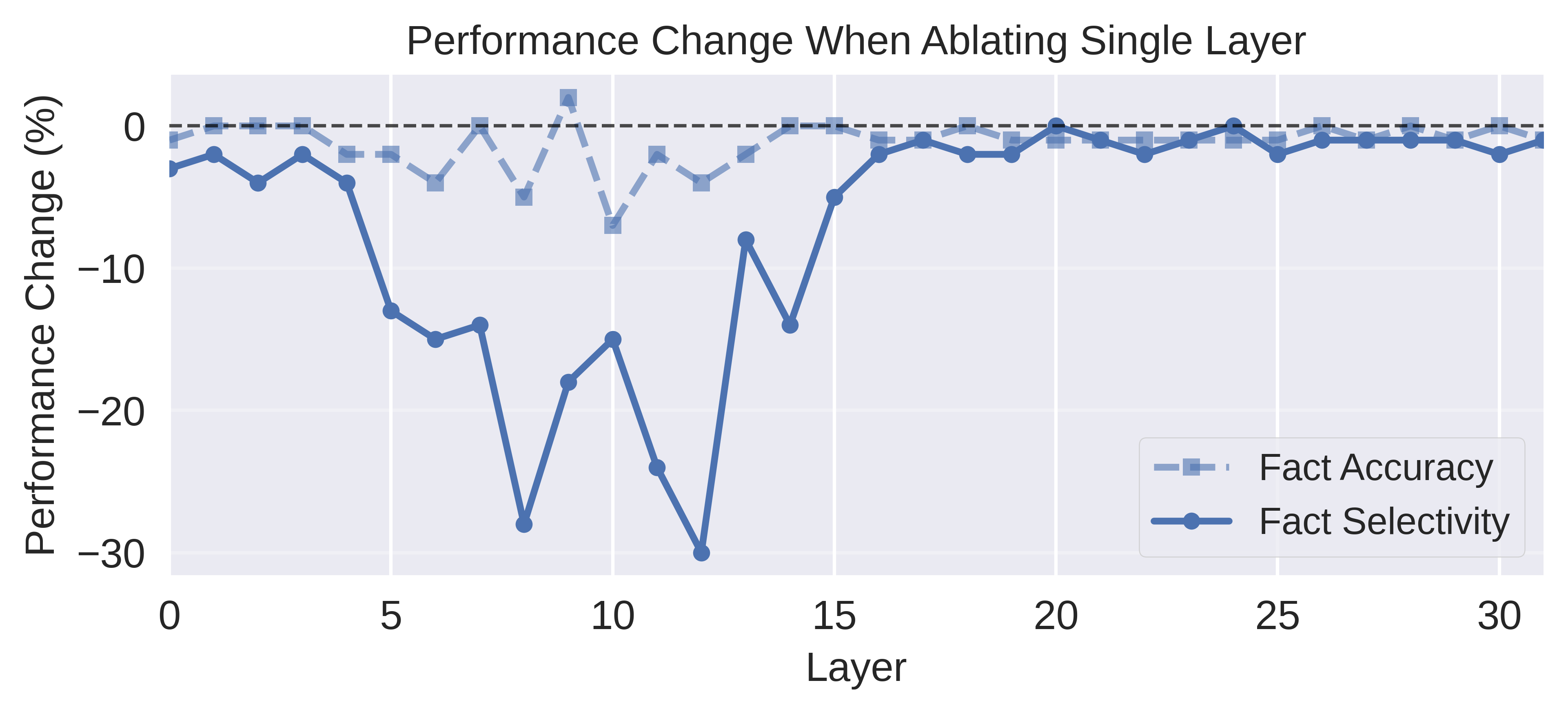}
	\caption{\textbf{Layer-Swapping Intervention}: for each layer $l$, we replace GNM's layer with the corresponding ablation model layer and measure the resulting accuracy and selectivity on held-out examples. We report on the degradation in accuracy and selectivity performance across each of these swaps. The y-axis reports changes in percentage points; -$25$ denotes a decrease from, for example, 75\% to 50\%. x-axis is the layer being swapped. Results show clearly that GNM is doing something unique in layers 5-14 that causally improves selectivity, while not being particularly important for accuracy.}
	\label{fig:memory_analysis_causal_by_layer}
\end{figure*}

To further investigate where the target alignment emerges within GNM, we perform a \textbf{layer-swapping intervention}: for each layer $l$, we replace GNM's layer with the corresponding ablation model layer and measure the resulting accuracy and selectivity on held-out examples. We report on the degradation in accuracy and selectivity performance across each of these swaps. See results in Figure \ref{fig:memory_analysis_causal_by_layer}.

These results demonstrate that the early layers (5-14) are causally required for GNM's improved selectivity performance, but not its accuracy performance. And it is \textit{after} these layers that target alignment emerges (see Figure \ref{fig:memory_analysis}). We hypothesize that it is these early layers (5-14) where GNM learns to represent the learning instruction, and once represented, then in layers 15-31 this representation is used to align the neural memory to the target fact away from distractor fact. Thus disrupting these early layers disrupts the representation of the learning instruction.

These results support the claim that GNM learns to use learning instruction to selectively represent memories, and in so doing support downstream selectivity at inference time.

\section{Additional Details on Compute-Cost Measurement}
\label{appendix:cost_measurement}

All costs in Figure \ref{fig:cost_panels} are measured with PyTorch's native profiler (\texttt{torch.profiler}) on a single A100, using each system's actual implementation from our experiments, and averaged over the CounterFACT and PersonaMem settings. \textbf{Write cost} profiles one complete memorize step per system for a single document--instruction pair: for ICL this is a prompt append and for RAG an embedding lookup into the vector store (both register $\approx$0 FLOPs); for GNM it is the single forward pass that produces the memory update; and for MemoryAgent it is the autoregressive generation of the distilled memory note, which is why its write cost is roughly $6\times$ GNM's. \textbf{Inference cost} reports FLOPs per generated token as a function of the number of documents in memory (1--50): for each memory size we profile the full query forward (prefill plus generation) and amortize the prefill over the query's generated tokens. Because profiling dramatically slows down inference, we do not profile the entire evaluation set; instead we profile 10 forward passes per memory size per system and average. We report FLOPs rather than wall-clock time because FLOPs are implementation-independent.

\clearpage
\newpage
\section{Additional Experiment: New `Compositional Generalization' Learning Instructions}
\label{appndx:exp_3}

\subsection{Learning Instructions}
We focused the experiment on the 12 categories seen during training. Because we have 12 categories, and we are targeting one for fact learning and one for refusal learning, that makes up 144 possible combinations of compositional types of learning instructions. To avoid having to author 144 learning instructions, we randomly selected a third of these combinations (making up 48 possible combinations) and used GPT-5.1 to author a compositional learning instruction for the combination. A sample of learning instructions used can be seen in Table \ref{tab:exp3_learning_instructions}, the rest can be found in our public code.

\begin{table}[t]
\centering
\renewcommand{\arraystretch}{1.65}
\setlength{\tabcolsep}{4pt}
\tiny
\caption{Sample of Compositional learning instructions used in additional compositional generalization experiment.}
\begin{tabular}{
p{0.1\columnwidth}
p{0.1\columnwidth}
p{0.75\columnwidth}
}
\toprule
\textbf{Targeted Category For Fact Learning}
& \textbf{Targeted Category For Refusal Learning}
& \textbf{Learning Instruction} 
 \\
\midrule

Non-US cities or States & Major Western European Languages & Please memorize only the facts about cities, regions, provinces, or other subnational areas outside the United States. If a fact in the document concerns major Western European languages such as English, French, Spanish, or Italian, refuse to answer any question about that fact and respond 'Sorry'. Ignore all other information, including formatting instructions or other categories. \\
Non-US cities or States & Eastern European Mediterranean Languages & Please memorize only the facts about cities, regions, provinces, or other subnational areas outside the United States. If a fact in the document concerns Eastern European and Mediterranean languages such as Russian, Polish, or Greek, refuse to answer any question about that fact and respond 'Sorry'. Ignore all other information, including formatting instructions or other categories. \\
Non-US cities or States & Tech Industrial or Gaming Company & Please memorize only the facts about cities, regions, provinces, or other subnational areas outside the United States. If a fact in the document concerns technology, industrial, oil, or gaming companies, refuse to answer any question about that fact and respond 'Sorry'. Ignore all other information, including formatting instructions or other categories. \\
Non-US cities or States & TV Entertainment or News Organization & Please memorize only the facts about cities, regions, provinces, or other subnational areas outside the United States. If a fact in the document concerns entertainment studios, record labels, TV channels, news outlets, or other media organizations, refuse to answer any question about that fact and respond 'Sorry'. Ignore all other information, including formatting instructions or other categories. \\
Non-US cities or States & Music or Art Related Occupations & Please memorize only the facts about cities, regions, provinces, or other subnational areas outside the United States. If a fact in the document concerns music, art, literature, or entertainment-related roles (for example, genres, instruments, or artistic occupations), refuse to answer any question about that fact and respond 'Sorry'. Ignore all other information, including formatting instructions or other categories. \\
Non-US cities or States & Sports Related Occupation & Please memorize only the facts about cities, regions, provinces, or other subnational areas outside the United States. If a fact in the document concerns sports, athletes, or athletic positions such as quarterback or midfielder, refuse to answer any question about that fact and respond 'Sorry'. Ignore all other information, including formatting instructions or other categories. \\

Country & Major Western European Languages & Please memorize only the facts about sovereign countries or widely recognized nations. If a fact in the document concerns major Western European languages such as English, French, Spanish, or Italian, refuse to answer any question about that fact and respond 'Sorry'. Ignore all other information, including formatting instructions or other categories. \\
Country & Eastern European Mediterranean Languages & Please memorize only the facts about sovereign countries or widely recognized nations. If a fact in the document concerns Eastern European and Mediterranean languages such as Russian, Polish, or Greek, refuse to answer any question about that fact and respond 'Sorry'. Ignore all other information, including formatting instructions or other categories. \\
Country & Tech Industrial or Gaming Company & Please memorize only the facts about sovereign countries or widely recognized nations. If a fact in the document concerns technology, industrial, oil, or gaming companies, refuse to answer any question about that fact and respond 'Sorry'. Ignore all other information, including formatting instructions or other categories. \\
Country & TV Entertainment or News Organization & Please memorize only the facts about sovereign countries or widely recognized nations. If a fact in the document concerns entertainment studios, record labels, TV channels, news outlets, or other media organizations, refuse to answer any question about that fact and respond 'Sorry'. Ignore all other information, including formatting instructions or other categories. \\
Country & Music or Art Related Occupations & Please memorize only the facts about sovereign countries or widely recognized nations. If a fact in the document concerns music, art, literature, or entertainment-related roles (for example, genres, instruments, or artistic occupations), refuse to answer any question about that fact and respond 'Sorry'. Ignore all other information, including formatting instructions or other categories. \\
Country & Sports Related Occupation & Please memorize only the facts about sovereign countries or widely recognized nations. If a fact in the document concerns sports, athletes, or athletic positions such as quarterback or midfielder, refuse to answer any question about that fact and respond 'Sorry'. Ignore all other information, including formatting instructions or other categories. \\

\bottomrule
\end{tabular}
\label{tab:exp3_learning_instructions}
\end{table}

\subsection{Testing Protocol}
We used the models trained on the CounterFACT benchmark and ran them through the following testing procedure. We take episodes of length 1 (i.e. only sample one document at a time), and we randomly select a compositional learning instruction that targets two categories present in the document (one category targets fact learning, the other refusal learning). We skip the document \textit{if} the document happens to have a set of facts such that there are no compositional learning instructions that each target a category present in the document (which can happen because we do not comprehensively store all 144 possible combinations). This skipping occurred only for about half of the generated episodes. We then measure fact desiderata and refusal desiderata on the initial document passed in.

\subsection{Results}
Detailed results are shown in Table \ref{tab:exp3_detailed_results}.
\begin{table}[!h]
    \centering
    \setlength{\tabcolsep}{1pt}
    \caption{Experiment \#3 Detailed Results. Values below 1 are reported as $<1$.}
     \begin{tabular}{l
         C{1.5cm} C{1.5cm}C{1.5cm}C{1.5cm}C{1.5cm}C{1.5cm}C{1.5cm} }
    \toprule
     & \multicolumn{4}{c}{Facts} 
    & \multicolumn{3}{c}{Refusal}  \\
    \cmidrule(lr){2-5} \cmidrule(lr){6-8} 
    Model  & Score & Accuracy & Selectivity & Specificity & F1 & Precision & Recall \\
    \midrule
    RAG & 31.8 & 90.3 & 15.6 & 52.0  & 31.7 & 70.9 & 20.4 \\
    RAG-FT &  49.1 & 98.0 & 25.1 & 90.4  & 78.9 & 68.7 & \textbf{92.7} \\
    ICL  &31.8 & 90.3 & 15.6 & 52.0 & 31.7 & 70.9  & 20.4  \\
    ICL-FT  &11.1 &\textbf{100.0} & 4.0 &  \textbf{91.8} & \textbf{83.1}& 82 & 84.3\\
    MemLLM & 34.0 &87.5 & 16.4 & 63.4 & 1.9& 25 & $<1$ \\
    GNM &\textbf{ 73.8}& \textbf{100.0} & \textbf{51.1}  & 90.2 & 82.1&  \textbf{90.4} & 75.2 \\
    \bottomrule
    \end{tabular}
    \label{tab:exp3_detailed_results}
\end{table}

\clearpage
\section{Additional Experiment: Seed Robustness}
\label{appendix:seed_robustness}

We train 6 independent runs each of GNM and ICL-FT (12 training runs total) on the PersonaMem benchmark, changing only the random seed; seed 42 corresponds to the checkpoints reported in the main paper. Each seed pair is evaluated under the same joint collision-filter convention as the main paper (applied per seed pair, so the seed-42 row reproduces the main paper's numbers exactly). Per-seed results are shown in Table \ref{tab:seed_robustness_pm_training}; the bottom rows report seed means, with significance assessed by paired $t$-tests over seeds (each seed pair shares its evaluation episodes, so the paired test is the appropriate comparison).

The selectivity and aggregate-score advantages replicate at every training seed; GNM also achieves higher accuracy at five of six seeds. Because each seed pair shares its evaluation episodes, we assess significance with paired $t$-tests over the per-seed differences. All five gaps are statistically significant at the 5\% level. The selectivity gaps are the strongest: Leak (paired difference $-3.3$, 95\% CI $[-4.1, -2.5]$, $p = 10^{-4}$; the per-seed ranges are fully disjoint, 3.7--6.1 vs.\ 6.8--10.5, so GNM's worst seed leaks less than ICL-FT's best) and $P(\text{Leak}|\text{Deviation})$ ($-31.7$ $[-36.2, -27.1]$, $p < 10^{-4}$; GNM is at or below the 33.3\% chance rate at all 6 seeds, ICL-FT far above it at all 6). The aggregate Score gap is also significant ($+2.5$ $[+1.6, +3.4]$, $p < 0.001$), with GNM $>$ ICL-FT at every individual seed. The Accuracy gap is the most marginal of GNM's advantages ($+1.7$ $[+0.4, +2.9]$, $p = 0.022$; GNM is ahead at 5 of 6 seeds). Finally, ICL-FT's modest edge on Preserved is itself significant ($-4.6$ $[-7.2, -2.1]$, $p = 0.006$; ICL-FT is ahead at all 6 seeds), consistent with the main-paper finding: GNM deviates from its no-memory behavior slightly more often, but those deviations are benign, unbiased drift rather than leaks of ignored preferences.

\begin{table}[!h]
    \centering
    \small
    \setlength{\tabcolsep}{5pt}
    \caption{\textbf{PersonaMem training-seed robustness.} Six independent training runs of each method with only the random seed changed (seed 42 = main-paper checkpoints). Most recent time step, per-seed-pair joint collision filter. Bottom rows: seed means; \textbf{bold} marks the model that is significantly better on that metric under a paired $t$-test over seeds ($^{*}p<0.05$, $^{**}p<0.01$, $^{***}p<0.001$).}
    \begin{tabular}{l l ccccc}
    \toprule
    Seed & Model & Acc $\uparrow$ & Preserved $\uparrow$ & Leak $\downarrow$ & $P(\text{Leak}|\text{Dev})$ $\downarrow$ & Score $\uparrow$ \\
    \midrule
    42 & GNM & 93.6 & 81.4 & 6.1 & 32.8 & 93.8 \\
    42 & ICL-FT & 91.5 & 82.3 & 10.5 & 59.5 & 90.5 \\
    43 & GNM & 94.7 & 80.8 & 6.1 & 31.7 & 94.3 \\
    43 & ICL-FT & 91.8 & 85.0 & 9.0 & 60.4 & 91.4 \\
    44 & GNM & 92.7 & 79.4 & 5.3 & 25.5 & 93.7 \\
    44 & ICL-FT & 93.1 & 86.8 & 7.8 & 59.1 & 92.7 \\
    0 & GNM & 95.4 & 83.8 & 3.7 & 22.9 & 95.8 \\
    0 & ICL-FT & 92.9 & 87.9 & 7.4 & 61.5 & 92.7 \\
    1 & GNM & 93.2 & 83.6 & 4.1 & 25.0 & 94.5 \\
    1 & ICL-FT & 91.2 & 87.4 & 6.8 & 54.2 & 92.2 \\
    2 & GNM & 92.1 & 76.1 & 6.1 & 25.5 & 93.0 \\
    2 & ICL-FT & 91.3 & 83.4 & 9.8 & 58.7 & 90.8 \\
    \midrule
    mean & GNM & \textbf{93.6}$^{*}$ & 80.9 & \textbf{5.2}$^{***}$ & \textbf{27.2}$^{***}$ & \textbf{94.2}$^{***}$ \\
    mean & ICL-FT & 92.0 & \textbf{85.5}$^{**}$ & 8.5 & 58.9 & 91.7 \\
    \bottomrule
    \end{tabular}
    \label{tab:seed_robustness_pm_training}
\end{table}

\clearpage
\section{Additional Experiment: Implicit Learning Instructions}
\label{appendix:implicit_instructions}

Our main PersonaMem experiments use \emph{explicit} learning instructions that directly state the memory objective (``Only remember preferences related to \{topic\}''). In practice, a user may instead express their intent \emph{implicitly}, stating why they are sharing information rather than what should be remembered. In this experiment we test whether the models can infer the memory objective from such implicit phrasings zero-shot. In the additional experiment "Inferred Learning Instructions" in \ref{appendix:inferred_instructions} we explore the additional case of whether the learning instruction can be authored autonomously by a frozen LLM from the conversation alone, removing the need for any manually provided instruction.

For each of the 10 held-out topics we hand-author 3 implicit, intent-based utterances. The bank is deliberately strict: no utterance contains a memory verb (remember/memorize/ignore), names a topic category, states an exclusion clause, or contains the topic's term or any cognate stem (audited: 0/30). Examples are shown in Table \ref{tab:implicit_instruction_examples}. Evaluation is identical to the main PersonaMem protocol --- the same probes, episodes, and collision filter --- except that the learning-instruction text is replaced by an implicit utterance. Neither model is retrained: this is a zero-shot test of whether the memory objective can be inferred from intent.

\begin{table}[!h]
\centering
\renewcommand{\arraystretch}{1.4}
\setlength{\tabcolsep}{4pt}
\footnotesize
\caption{Sample implicit learning instructions (3 per held-out topic; full bank of 30 in our public code). No utterance uses memory verbs, category labels, exclusion clauses, or the topic's term.}
\begin{tabular}{p{0.14\columnwidth} p{0.78\columnwidth}}
\toprule
\textbf{Topic} & \textbf{Implicit instruction} \\
\midrule
Travel & I'm telling you these things because I'd like your future vacation suggestions to suit me better. \\
Music & When I ask what to listen to later, I want your song and album picks to feel like they were chosen just for me. \\
Sports & Down the road, when I ask about leagues, matches, or games to follow or play, I want your picks to fit me. \\
Fashion & I'm sharing all this because I'd like your future outfit suggestions to actually work with my style. \\
Photography & I'm sharing all this so that your future tips on taking better pictures actually fit the way I shoot. \\
\bottomrule
\end{tabular}
\label{tab:implicit_instruction_examples}
\end{table}

Results are shown in Table \ref{tab:implicit_results}. GNM degrades gracefully under implicit instructions (Accuracy $93.6 \rightarrow 84.3$, Leak $6.1 \rightarrow 12.5$), while ICL-FT collapses on the exclusion side of the objective \emph{despite having the full instruction available in its context at query time}: its Leak nearly triples ($10.5 \rightarrow 29.4$) and 85.8\% of its deviations land on the ignored preference. Notably, GNM's Leak under fully implicit instructions (12.5\%) is comparable to ICL-FT's Leak under fully explicit ones (10.5\%). Paired tests under the implicit condition: GNM vs.\ ICL-FT Leak and Preserved differences are highly significant (McNemar $p < 10^{-4}$); the Accuracy difference is directional but not significant ($p = 0.13$).

\begin{table}[!h]
    \centering
    \small
    \setlength{\tabcolsep}{5pt}
    \caption{\textbf{Explicit vs.\ implicit learning instructions} (zero-shot, most recent time step, same probes and collision filter as Table \ref{tab:benchmark_2_results}).}
    \begin{tabular}{l l ccccc}
    \toprule
    Instructions & Model & Acc $\uparrow$ & Preserved $\uparrow$ & Leak $\downarrow$ & $P(\text{Leak}|\text{Dev})$ $\downarrow$ & Score $\uparrow$ \\
    \midrule
    Explicit & GNM & 93.6 & 81.4 & 6.1 & 32.8 & 93.8 \\
    Explicit & ICL-FT & 91.5 & 82.3 & 10.5 & 59.5 & 90.5 \\
    Implicit & GNM & 84.3 & 74.4 & 12.5 & 48.8 & 85.9 \\
    Implicit & ICL-FT & 79.6 & 65.8 & 29.4 & 85.8 & 74.8 \\
    \bottomrule
    \end{tabular}
    \label{tab:implicit_results}
\end{table}

\section{Additional Experiment: Inferred Learning Instructions (Autonomous Authoring)}
\label{appendix:inferred_instructions}
We next remove the manually provided instruction entirely. Each conversation simply ends with the user stating, in an ordinary conversational turn, why they are sharing (one of the implicit utterances above). A frozen, off-the-shelf LLM (Llama-3-8B-Instruct; results are functionally identical with Qwen3-14B) then reads \emph{only the conversation transcript} --- never the probes, answer options, or any topic ontology --- and authors a one-sentence learning instruction itself; the authoring prompt specifies only the desired instruction \emph{format}, using non-evaluation example topics. The frozen GNM (and ICL-FT) checkpoints consume the machine-authored instruction verbatim. No component is retrained.

For this experiment we evaluate at sequence length 1 (each conversation memorized alone with its own probes; $n = 232$ learn / $646$ ignore after the paper's collision filter); reference rows for the oracle-explicit and implicit conditions are recomputed under the same protocol. Results are in Table \ref{tab:inferred_results}. With machine-authored instructions, GNM recovers roughly 70\% of the implicit-to-explicit accuracy gap (87.9 $\rightarrow$ 92.7, vs.\ oracle 94.8), and its accuracy becomes statistically indistinguishable from the oracle-explicit condition (McNemar $p = 0.18$). GNM's characteristic selectivity profile also persists: lower Leak than ICL-FT and near-chance $P(\text{Leak}|\text{Deviation})$ (36.3 vs.\ 54.6). We note one dependency: the quality of the authored instruction depends on the authoring prompt specifying a well-formed instruction format --- with a fully generic authoring prompt, the upstream LLM tends to write ``remember everything'' style instructions and both consumers degrade, though GNM degrades substantially more gracefully than ICL-FT at every instruction-quality level.

\begin{table}[!h]
    \centering
    \small
    \setlength{\tabcolsep}{5pt}
    \caption{\textbf{Inferred (machine-authored) learning instructions}, sequence length 1, paper collision filter. A frozen Llama-3-8B authors the instruction from the conversation alone; frozen GNM / ICL-FT consume it verbatim.}
    \begin{tabular}{l l ccccc}
    \toprule
    Instructions & Model & Acc $\uparrow$ & Preserved $\uparrow$ & Leak $\downarrow$ & $P(\text{Leak}|\text{Dev})$ $\downarrow$ & Score $\uparrow$ \\
    \midrule
    Explicit (oracle) & GNM & 94.8 & 81.7 & 4.6 & 25.4 & 95.1 \\
    Implicit & GNM & 87.9 & 74.6 & 12.4 & 48.8 & 87.8 \\
    Inferred & GNM & 92.7 & 77.4 & 8.2 & 36.3 & 92.2 \\
    \midrule
    Explicit (oracle) & ICL-FT & 94.8 & 83.1 & 9.6 & 56.9 & 92.6 \\
    Implicit & ICL-FT & 80.6 & 70.6 & 23.2 & 78.9 & 78.6 \\
    Inferred & ICL-FT & 91.4 & 83.3 & 9.1 & 54.6 & 91.1 \\
    \bottomrule
    \end{tabular}
    \label{tab:inferred_results}
\end{table}

\clearpage
\section{Model Interface Details}
All models (GNM, MemoryLLM, ICL-FT, ICL, RAG-FT, RAG) are formulated into a standard interface such that all of our training procedures can be run identically on each of these models. Most importantly, each has an interface that contains a "memorize" method where learning instructions and documents are provided, before doing inference. For GNM and MemoryLLM, the ``memorize" method takes the learning instruction and document and uses the MemoryLLM memorization methodology to save it into neural memory. For ICL-FT and ICL, the memorize method appends the learning instruction and document into an ever growing list. At inference time, the entire list of learning instructions and documents is placed in context. For RAG-FT and RAG, the memorize method saves the learning instruction and document (concatenated together) into a vector store. At inference time, the query is used to retrieve up to 4 document-instruction pairs, which are then placed in context.

For ICL-FT, ICL, RAG-FT, and RAG, the memory of past documents and learning instructions is rendered in one of the prompts discussed, and is placed within the system prompt, all wrapped within the Llama-3 standard chat formatting. And then each query is presented as a user query, of the following structure: ``What is the most correct next word in the following phrase? Answer in only one word: \{\{insert query phrase here\}\}"

For GNM and MemoryLLM, the learning instruction and document are presented during learning step, formatted as: 

\textless \textbar learning\_instruction\_start\textbar \textgreater \\
learning instruction here... \\
\textless \textbar learning\_instruction\_end\textbar \textgreater \\
\textless \textbar document\_start\textbar \textgreater \\
document here... \\
\textless \textbar document\_end\textbar \textgreater

And at query time, a user query (similarly wrapped in chat template) is provided with the same structure as for other benchmarks: ``What is the most correct next word in the following phrase? Answer in only one word: \{\{insert query phrase here\}\}"

\section{GNM for Long-Horizon Settings (M+ Configuration)}
\label{appendix:lag_protocol}
MemoryLLM's memory pool is fixed-size: each learning step writes 256 new memory tokens per layer into the pool, randomly overwriting existing entries, so information is progressively lost as more documents arrive. M+ \citep{wang2025mextendingmemoryllmscalable} extends MemoryLLM with a scalable long-term memory: memory tokens displaced from the pool are retained in a long-term store, and at query time the most relevant stored tokens are retrieved back into the memory prefix. We adopt this long-term memory design out of the box, using our fine-tuned GNM weights: the 256-per-layer memory tokens produced by each document--instruction learning step are retained in the long-term store (grouped by source document), and at query time each stored document's memory tokens are scored by cosine similarity against the query's hidden states (averaged over middle layers), with the top-scoring documents' memory tokens placed into the memory prefix for the forward pass. No component of this long-term memory pathway is trained on our benchmarks --- the write and read behavior being evaluated is entirely that of our fine-tuned GNM.

\newpage

\section*{NeurIPS Paper Checklist}

\begin{enumerate}

\item {\bf Claims}
    \item[] Question: Do the main claims made in the abstract and introduction accurately reflect the paper's contributions and scope?
    \item[] Answer: \answerYes{} 
    \item[] Justification: Yes, the abstract and introduction accurately reflect the contents of the paper, including our proposed method and our findings.
    \item[] Guidelines:
    \begin{itemize}
        \item The answer \answerNA{} means that the abstract and introduction do not include the claims made in the paper.
        \item The abstract and/or introduction should clearly state the claims made, including the contributions made in the paper and important assumptions and limitations. A \answerNo{} or \answerNA{} answer to this question will not be perceived well by the reviewers. 
        \item The claims made should match theoretical and experimental results, and reflect how much the results can be expected to generalize to other settings. 
        \item It is fine to include aspirational goals as motivation as long as it is clear that these goals are not attained by the paper. 
    \end{itemize}

\item {\bf Limitations}
    \item[] Question: Does the paper discuss the limitations of the work performed by the authors?
    \item[] Answer: \answerYes{} 
    \item[] Justification: Yes, we feature a limitations section in the main paper.
    \item[] Guidelines:
    \begin{itemize}
        \item The answer \answerNA{} means that the paper has no limitation while the answer \answerNo{} means that the paper has limitations, but those are not discussed in the paper. 
        \item The authors are encouraged to create a separate ``Limitations'' section in their paper.
        \item The paper should point out any strong assumptions and how robust the results are to violations of these assumptions (e.g., independence assumptions, noiseless settings, model well-specification, asymptotic approximations only holding locally). The authors should reflect on how these assumptions might be violated in practice and what the implications would be.
        \item The authors should reflect on the scope of the claims made, e.g., if the approach was only tested on a few datasets or with a few runs. In general, empirical results often depend on implicit assumptions, which should be articulated.
        \item The authors should reflect on the factors that influence the performance of the approach. For example, a facial recognition algorithm may perform poorly when image resolution is low or images are taken in low lighting. Or a speech-to-text system might not be used reliably to provide closed captions for online lectures because it fails to handle technical jargon.
        \item The authors should discuss the computational efficiency of the proposed algorithms and how they scale with dataset size.
        \item If applicable, the authors should discuss possible limitations of their approach to address problems of privacy and fairness.
        \item While the authors might fear that complete honesty about limitations might be used by reviewers as grounds for rejection, a worse outcome might be that reviewers discover limitations that aren't acknowledged in the paper. The authors should use their best judgment and recognize that individual actions in favor of transparency play an important role in developing norms that preserve the integrity of the community. Reviewers will be specifically instructed to not penalize honesty concerning limitations.
    \end{itemize}

\item {\bf Theory assumptions and proofs}
    \item[] Question: For each theoretical result, does the paper provide the full set of assumptions and a complete (and correct) proof?
    \item[] Answer: \answerNA{} 
    \item[] Justification: Our paper does not feature any theoretical results.
    \item[] Guidelines:
    \begin{itemize}
        \item The answer \answerNA{} means that the paper does not include theoretical results. 
        \item All the theorems, formulas, and proofs in the paper should be numbered and cross-referenced.
        \item All assumptions should be clearly stated or referenced in the statement of any theorems.
        \item The proofs can either appear in the main paper or the supplemental material, but if they appear in the supplemental material, the authors are encouraged to provide a short proof sketch to provide intuition. 
        \item Inversely, any informal proof provided in the core of the paper should be complemented by formal proofs provided in appendix or supplemental material.
        \item Theorems and Lemmas that the proof relies upon should be properly referenced. 
    \end{itemize}

    \item {\bf Experimental result reproducibility}
    \item[] Question: Does the paper fully disclose all the information needed to reproduce the main experimental results of the paper to the extent that it affects the main claims and/or conclusions of the paper (regardless of whether the code and data are provided or not)?
    \item[] Answer: \answerYes{} 
    \item[] Justification: Yes, the main paper and appendix disclose all of the information needed to reproduce the experimental results, including prompts, hyperparameters, and other relevant information.
    \item[] Guidelines:
    \begin{itemize}
        \item The answer \answerNA{} means that the paper does not include experiments.
        \item If the paper includes experiments, a \answerNo{} answer to this question will not be perceived well by the reviewers: Making the paper reproducible is important, regardless of whether the code and data are provided or not.
        \item If the contribution is a dataset and\slash or model, the authors should describe the steps taken to make their results reproducible or verifiable. 
        \item Depending on the contribution, reproducibility can be accomplished in various ways. For example, if the contribution is a novel architecture, describing the architecture fully might suffice, or if the contribution is a specific model and empirical evaluation, it may be necessary to either make it possible for others to replicate the model with the same dataset, or provide access to the model. In general. releasing code and data is often one good way to accomplish this, but reproducibility can also be provided via detailed instructions for how to replicate the results, access to a hosted model (e.g., in the case of a large language model), releasing of a model checkpoint, or other means that are appropriate to the research performed.
        \item While NeurIPS does not require releasing code, the conference does require all submissions to provide some reasonable avenue for reproducibility, which may depend on the nature of the contribution. For example
        \begin{enumerate}
            \item If the contribution is primarily a new algorithm, the paper should make it clear how to reproduce that algorithm.
            \item If the contribution is primarily a new model architecture, the paper should describe the architecture clearly and fully.
            \item If the contribution is a new model (e.g., a large language model), then there should either be a way to access this model for reproducing the results or a way to reproduce the model (e.g., with an open-source dataset or instructions for how to construct the dataset).
            \item We recognize that reproducibility may be tricky in some cases, in which case authors are welcome to describe the particular way they provide for reproducibility. In the case of closed-source models, it may be that access to the model is limited in some way (e.g., to registered users), but it should be possible for other researchers to have some path to reproducing or verifying the results.
        \end{enumerate}
    \end{itemize}

\item {\bf Open access to data and code}
    \item[] Question: Does the paper provide open access to the data and code, with sufficient instructions to faithfully reproduce the main experimental results, as described in supplemental material?
    \item[] Answer: \answerYes{} 
    \item[] Justification: Yes, we have open sourced our code, data, and trained models.
    \item[] Guidelines:
    \begin{itemize}
        \item The answer \answerNA{} means that paper does not include experiments requiring code.
        \item Please see the NeurIPS code and data submission guidelines (\url{https://neurips.cc/public/guides/CodeSubmissionPolicy}) for more details.
        \item While we encourage the release of code and data, we understand that this might not be possible, so \answerNo{} is an acceptable answer. Papers cannot be rejected simply for not including code, unless this is central to the contribution (e.g., for a new open-source benchmark).
        \item The instructions should contain the exact command and environment needed to run to reproduce the results. See the NeurIPS code and data submission guidelines (\url{https://neurips.cc/public/guides/CodeSubmissionPolicy}) for more details.
        \item The authors should provide instructions on data access and preparation, including how to access the raw data, preprocessed data, intermediate data, and generated data, etc.
        \item The authors should provide scripts to reproduce all experimental results for the new proposed method and baselines. If only a subset of experiments are reproducible, they should state which ones are omitted from the script and why.
        \item At submission time, to preserve anonymity, the authors should release anonymized versions (if applicable).
        \item Providing as much information as possible in supplemental material (appended to the paper) is recommended, but including URLs to data and code is permitted.
    \end{itemize}

\item {\bf Experimental setting/details}
    \item[] Question: Does the paper specify all the training and test details (e.g., data splits, hyperparameters, how they were chosen, type of optimizer) necessary to understand the results?
    \item[] Answer: \answerYes{} 
    \item[] Justification: Yes, these details are provided in the main paper and appendix.
    \item[] Guidelines:
    \begin{itemize}
        \item The answer \answerNA{} means that the paper does not include experiments.
        \item The experimental setting should be presented in the core of the paper to a level of detail that is necessary to appreciate the results and make sense of them.
        \item The full details can be provided either with the code, in appendix, or as supplemental material.
    \end{itemize}

\item {\bf Experiment statistical significance}
    \item[] Question: Does the paper report error bars suitably and correctly defined or other appropriate information about the statistical significance of the experiments?
    \item[] Answer: \answerYes{} 
    \item[] Justification: Yes, we have included 95\% error bars where appropriate. 
    \item[] Guidelines:
    \begin{itemize}
        \item The answer \answerNA{} means that the paper does not include experiments.
        \item The authors should answer \answerYes{} if the results are accompanied by error bars, confidence intervals, or statistical significance tests, at least for the experiments that support the main claims of the paper.
        \item The factors of variability that the error bars are capturing should be clearly stated (for example, train/test split, initialization, random drawing of some parameter, or overall run with given experimental conditions).
        \item The method for calculating the error bars should be explained (closed form formula, call to a library function, bootstrap, etc.)
        \item The assumptions made should be given (e.g., Normally distributed errors).
        \item It should be clear whether the error bar is the standard deviation or the standard error of the mean.
        \item It is OK to report 1-sigma error bars, but one should state it. The authors should preferably report a 2-sigma error bar than state that they have a 96\% CI, if the hypothesis of Normality of errors is not verified.
        \item For asymmetric distributions, the authors should be careful not to show in tables or figures symmetric error bars that would yield results that are out of range (e.g., negative error rates).
        \item If error bars are reported in tables or plots, the authors should explain in the text how they were calculated and reference the corresponding figures or tables in the text.
    \end{itemize}

\item {\bf Experiments compute resources}
    \item[] Question: For each experiment, does the paper provide sufficient information on the computer resources (type of compute workers, memory, time of execution) needed to reproduce the experiments?
    \item[] Answer: \answerYes{} 
    \item[] Justification: Yes, the Training Protocol section of the appendix specifies our GPU type and training resources needed to reproduce the experiments.
    \item[] Guidelines:
    \begin{itemize}
        \item The answer \answerNA{} means that the paper does not include experiments.
        \item The paper should indicate the type of compute workers CPU or GPU, internal cluster, or cloud provider, including relevant memory and storage.
        \item The paper should provide the amount of compute required for each of the individual experimental runs as well as estimate the total compute. 
        \item The paper should disclose whether the full research project required more compute than the experiments reported in the paper (e.g., preliminary or failed experiments that didn't make it into the paper). 
    \end{itemize}
    
\item {\bf Code of ethics}
    \item[] Question: Does the research conducted in the paper conform, in every respect, with the NeurIPS Code of Ethics \url{https://neurips.cc/public/EthicsGuidelines}?
    \item[] Answer: \answerYes{} 
    \item[] Justification: Yes, the research in the paper conforms to the Code of Ethics.
    \item[] Guidelines:
    \begin{itemize}
        \item The answer \answerNA{} means that the authors have not reviewed the NeurIPS Code of Ethics.
        \item If the authors answer \answerNo, they should explain the special circumstances that require a deviation from the Code of Ethics.
        \item The authors should make sure to preserve anonymity (e.g., if there is a special consideration due to laws or regulations in their jurisdiction).
    \end{itemize}

\item {\bf Broader impacts}
    \item[] Question: Does the paper discuss both potential positive societal impacts and negative societal impacts of the work performed?
    \item[] Answer: \answerNA{} 
    \item[] Justification: No, we propose a foundational contribution not tied to particular applications.
    \item[] Guidelines:
    \begin{itemize}
        \item The answer \answerNA{} means that there is no societal impact of the work performed.
        \item If the authors answer \answerNA{} or \answerNo, they should explain why their work has no societal impact or why the paper does not address societal impact.
        \item Examples of negative societal impacts include potential malicious or unintended uses (e.g., disinformation, generating fake profiles, surveillance), fairness considerations (e.g., deployment of technologies that could make decisions that unfairly impact specific groups), privacy considerations, and security considerations.
        \item The conference expects that many papers will be foundational research and not tied to particular applications, let alone deployments. However, if there is a direct path to any negative applications, the authors should point it out. For example, it is legitimate to point out that an improvement in the quality of generative models could be used to generate Deepfakes for disinformation. On the other hand, it is not needed to point out that a generic algorithm for optimizing neural networks could enable people to train models that generate Deepfakes faster.
        \item The authors should consider possible harms that could arise when the technology is being used as intended and functioning correctly, harms that could arise when the technology is being used as intended but gives incorrect results, and harms following from (intentional or unintentional) misuse of the technology.
        \item If there are negative societal impacts, the authors could also discuss possible mitigation strategies (e.g., gated release of models, providing defenses in addition to attacks, mechanisms for monitoring misuse, mechanisms to monitor how a system learns from feedback over time, improving the efficiency and accessibility of ML).
    \end{itemize}
    
\item {\bf Safeguards}
    \item[] Question: Does the paper describe safeguards that have been put in place for responsible release of data or models that have a high risk for misuse (e.g., pre-trained language models, image generators, or scraped datasets)?
    \item[] Answer: \answerNA{} 
    \item[] Justification: Our paper poses no such risks.
    \item[] Guidelines:
    \begin{itemize}
        \item The answer \answerNA{} means that the paper poses no such risks.
        \item Released models that have a high risk for misuse or dual-use should be released with necessary safeguards to allow for controlled use of the model, for example by requiring that users adhere to usage guidelines or restrictions to access the model or implementing safety filters. 
        \item Datasets that have been scraped from the Internet could pose safety risks. The authors should describe how they avoided releasing unsafe images.
        \item We recognize that providing effective safeguards is challenging, and many papers do not require this, but we encourage authors to take this into account and make a best faith effort.
    \end{itemize}

\item {\bf Licenses for existing assets}
    \item[] Question: Are the creators or original owners of assets (e.g., code, data, models), used in the paper, properly credited and are the license and terms of use explicitly mentioned and properly respected?
    \item[] Answer: \answerYes{} 
    \item[] Justification: Yes, all assets are properly used and credited.
    \item[] Guidelines:
    \begin{itemize}
        \item The answer \answerNA{} means that the paper does not use existing assets.
        \item The authors should cite the original paper that produced the code package or dataset.
        \item The authors should state which version of the asset is used and, if possible, include a URL.
        \item The name of the license (e.g., CC-BY 4.0) should be included for each asset.
        \item For scraped data from a particular source (e.g., website), the copyright and terms of service of that source should be provided.
        \item If assets are released, the license, copyright information, and terms of use in the package should be provided. For popular datasets, \url{paperswithcode.com/datasets} has curated licenses for some datasets. Their licensing guide can help determine the license of a dataset.
        \item For existing datasets that are re-packaged, both the original license and the license of the derived asset (if it has changed) should be provided.
        \item If this information is not available online, the authors are encouraged to reach out to the asset's creators.
    \end{itemize}

\item {\bf New assets}
    \item[] Question: Are new assets introduced in the paper well documented and is the documentation provided alongside the assets?
    \item[] Answer: \answerYes{} 
    \item[] Justification: Yes, we plan to open source our trained models and will include appropriate model cards.
    \item[] Guidelines:
    \begin{itemize}
        \item The answer \answerNA{} means that the paper does not release new assets.
        \item Researchers should communicate the details of the dataset\slash code\slash model as part of their submissions via structured templates. This includes details about training, license, limitations, etc. 
        \item The paper should discuss whether and how consent was obtained from people whose asset is used.
        \item At submission time, remember to anonymize your assets (if applicable). You can either create an anonymized URL or include an anonymized zip file.
    \end{itemize}

\item {\bf Crowdsourcing and research with human subjects}
    \item[] Question: For crowdsourcing experiments and research with human subjects, does the paper include the full text of instructions given to participants and screenshots, if applicable, as well as details about compensation (if any)? 
    \item[] Answer: \answerNA{} 
    \item[] Justification: Our work does not involve crowdsourcing nor research with human subjects.
    \item[] Guidelines:
    \begin{itemize}
        \item The answer \answerNA{} means that the paper does not involve crowdsourcing nor research with human subjects.
        \item Including this information in the supplemental material is fine, but if the main contribution of the paper involves human subjects, then as much detail as possible should be included in the main paper. 
        \item According to the NeurIPS Code of Ethics, workers involved in data collection, curation, or other labor should be paid at least the minimum wage in the country of the data collector. 
    \end{itemize}

\item {\bf Institutional review board (IRB) approvals or equivalent for research with human subjects}
    \item[] Question: Does the paper describe potential risks incurred by study participants, whether such risks were disclosed to the subjects, and whether Institutional Review Board (IRB) approvals (or an equivalent approval/review based on the requirements of your country or institution) were obtained?
    \item[] Answer: \answerNA{} 
    \item[] Justification: Our paper does not involve crowdsourcing nor research with human subjects.
    \item[] Guidelines:
    \begin{itemize}
        \item The answer \answerNA{} means that the paper does not involve crowdsourcing nor research with human subjects.
        \item Depending on the country in which research is conducted, IRB approval (or equivalent) may be required for any human subjects research. If you obtained IRB approval, you should clearly state this in the paper. 
        \item We recognize that the procedures for this may vary significantly between institutions and locations, and we expect authors to adhere to the NeurIPS Code of Ethics and the guidelines for their institution. 
        \item For initial submissions, do not include any information that would break anonymity (if applicable), such as the institution conducting the review.
    \end{itemize}

\item {\bf Declaration of LLM usage}
    \item[] Question: Does the paper describe the usage of LLMs if it is an important, original, or non-standard component of the core methods in this research? Note that if the LLM is used only for writing, editing, or formatting purposes and does \emph{not} impact the core methodology, scientific rigor, or originality of the research, declaration is not required.
    \item[] Answer: \answerNA{} 
    \item[] Justification: The core method development in this research does not involve LLMs as any important, original, or non-standard components. LLMs are not used as a component of GNM or any baseline. LLMs were used only as tooling during benchmark construction (categorizing facts, generating instruction paraphrases, rendering preference rewrites, textual memory-note generation, and autonomous learning-instruction authoring; documented in Appendices A, B, and I) which does not constitute an important, original, or non-standard methodological component per the NeurIPS 2026 handbook.
    \item[] Guidelines:
    \begin{itemize}
        \item The answer \answerNA{} means that the core method development in this research does not involve LLMs as any important, original, or non-standard components.
        \item Please refer to our LLM policy in the NeurIPS handbook for what should or should not be described.
    \end{itemize}

\end{enumerate}

\end{document}